\documentclass[pdflatex,iicol]{sn-jnl}

\usepackage{graphicx}%
\usepackage{bm}
 \usepackage{optidef}
\usepackage{multirow}%
\usepackage{amsmath,amssymb,amsfonts}%
\usepackage{amsthm}%
\usepackage{mathrsfs}%
\usepackage[title]{appendix}%
\usepackage{xcolor}%
\usepackage{textcomp}%
\usepackage{manyfoot}%
\usepackage{booktabs}%
\usepackage{algorithm}%
\usepackage{algorithmicx}%
\usepackage{algpseudocode}%
\usepackage{listings}%
\usepackage[small]{caption}%

\theoremstyle{thmstyleone}%
\theoremstyle{thmstyletwo}%

\theoremstyle{thmstylethree}%

\usepackage{subfig}

\usepackage[acronym]{glossaries}
\newacronym{acw}{ACW}{aerial co-worker}
\newacronym{am}{AM}{aerial manipulator}
\newacronym{ar}{AR}{aerial robot}
\newacronym{com}{CoM}{center of mass}
\newacronym{ee}{EE}{end effector}
\newacronym[longplural={degrees of freedom}]{dof}{DoF}{degrees of freedom}
\newacronym{genom}{GenoM}{Generator of Modules}
\newacronym{hri}{HRI}{Human-Robot Interaction}
\newacronym{imu}{IMU}{Inertial Measurement Unit}
\newacronym{matlab}{MATLAB}{MATrix LABoratory}
\newacronym{mocap}{MoCap}{motion capture}
\newacronym{mrav}{MRAV}{multi-rotor aerial vehicle}
\newacronym{ndt}{NDT}{non-destructive testing}
\newacronym{ne}{NE}{Newton-Euler}
\newacronym{phari}{pHARI}{physical human-aerial robot interaction}
\newacronym{phri}{pHRI}{physical human-robot interaction}
\newacronym{ros}{ROS}{Robot Operating System}
\newacronym{wrt}{w.r.t.}{with respect to}

\newacronym{od}{OD}{omnidirectional}
\newacronym{mdt}{MDT}{Multi-directional Thrust}
\newacronym{udt}{UDT}{Uni-directional Thrust}
\newacronym{oa}{OA}{over-actuated}
\newacronym{ua}{UA}{underactuated}
\newacronym{fa}{FA}{Fully-Actuated}

\usepackage{color}

\newcommand{\vect}[1]{\bm{#1}}		   
\newcommand{\matr}[1]{\mathbf{#1}}		       

\newcommand{\vE}[1]{\vect{e}_{#1}}			

\newcommand{\eye}[1]{\matr{I}_{#1}}

\newcommand{\pD}{\vect{p}^d}  
 
\newcommand{\ddpD}{\ddot{\vect{p}}^d}
\newcommand{\rotMatD}{{\mathbf{R}}^d}
\newcommand{\angVelD}{{\vect{\omega}}^d}
\newcommand{\angAccD}{{\dot{\vect{\omega}}}^d}
\newcommand{\errR}{{\vect{e}}_R}
\newcommand{\gainK}[1]{\matr{K}_{#1}}  
\newcommand{\pos}{\vect{p}}          
\newcommand{\angVel}{\vect{\omega}}   

\newcommand{\rot}{\mathbf{R}}          

\newcommand{\inertia}{\mathbf{M}}	
\newcommand{\gravityCoriolis}{\bm{\mu}} 

\newcommand{\rank}{\text{rank}}  

\usepackage[absolute]{textpos}

\textblockorigin{10mm}{10mm}
\begin{document}
\begin{textblock}{4.4}(0.9,0)	
	\textcolor{red}{This version of the article has been accepted for publication, after peer review. But is not the Version of Record and does not reflect post-acceptance improvements, or any corrections. The Version of Record of this article is published in Journal of Intelligent \& Robotic Systems and is available online at \href[]{https://doi.org/10.1007/s10846-024-02054-x}{https://doi.org/10.1007/s10846-024-02054-x}}
 \end{textblock}
\title[Article Title]{Modelling, Analysis,  and Control of OmniMorph: an Omnidirectional Morphing Multi-rotor UAV}


\author*[1]{\fnm{Youssef} \sur{Aboudorra}}\email{y.a.l.a.aboudorra@utwente.nl}

\author*[1]{\fnm{Chiara} \sur{Gabellieri}}\email{c.gabellieri@utwente.nl}

\author[1]{\fnm{Ralph} \sur{Brantjes}}\email{r.j.brantjes@student.utwente.nl}

\author[1]{\fnm{Quentin} \sur{Sablé}}\email{q.l.g.sable@utwente.nl}

\author[1,2]{\fnm{Antonio} \sur{Franchi}}\email{a.franchi@utwente.nl}
\affil*[1]{\orgdiv{Robotics and Mechatronics group, Faculty of Electrical Engineering, Mathematics \& Computer Science}, \orgname{University of Twente}, \orgaddress{ \city{Enschede}, \country{The Netherlands}}}

\affil[2]{\orgdiv{Department of Computer, Control and Management Engineering}, \orgname{Sapienza University of Rome}, \orgaddress{ \city{Rome}, \postcode{00185}, \country{Italy}}}




\abstract{	
  This paper introduces for the first time the design, modelling, and control of a novel morphing multi-rotor Unmanned Aerial Vehicle (UAV) that we call the OmniMorph. The morphing ability allows the selection of the configuration that optimizes energy consumption while ensuring the needed maneuverability for the required task. The most energy-efficient \textit{uni-directional thrust} (UDT) configuration can be used, e.g., during standard point-to-point displacements.
   \textit{Fully-actuated} (\acrshort{fa}) and \textit{omnidirectional}  (\acrshort{od}) configurations can be instead used for full pose tracking, such as, e.g., constant attitude horizontal motions and full rotations on the spot, and for full wrench 6D interaction control and 6D disturbance rejection. Morphing is obtained using a single servomotor, allowing possible minimization of weight, costs, and maintenance complexity.
   The actuation properties are studied, and an optimal controller that compromises between performance and control effort is proposed and validated in realistic simulations. \textcolor{black}{Preliminary tests on the prototype are presented to assess the propellers' mutual aerodynamic interference.}}
		
\maketitle	
	\section{Introduction}\label{sec:intro}
		Over the last decade, the design, perception, and control of Unmanned Aerial Vehicles (UAVs)  have substantially evolved in terms of functionalities and capabilities indoors or outdoors. Research now considers the use of UAVs as Aerial Robotic Manipulators that can interact physically with the environment \cite{ollero_past2022} and humans   \cite{tognon_physical2021};  contact-based inspection \cite{tognon2019truly}, object manipulation \cite{gabellieri2023equilibria}, and assisting human workers \cite{corsini2022nonlinear} are some of the typical tasks. 
  \begin{figure}[t]
	\centering
	\includegraphics[width=0.9\linewidth]{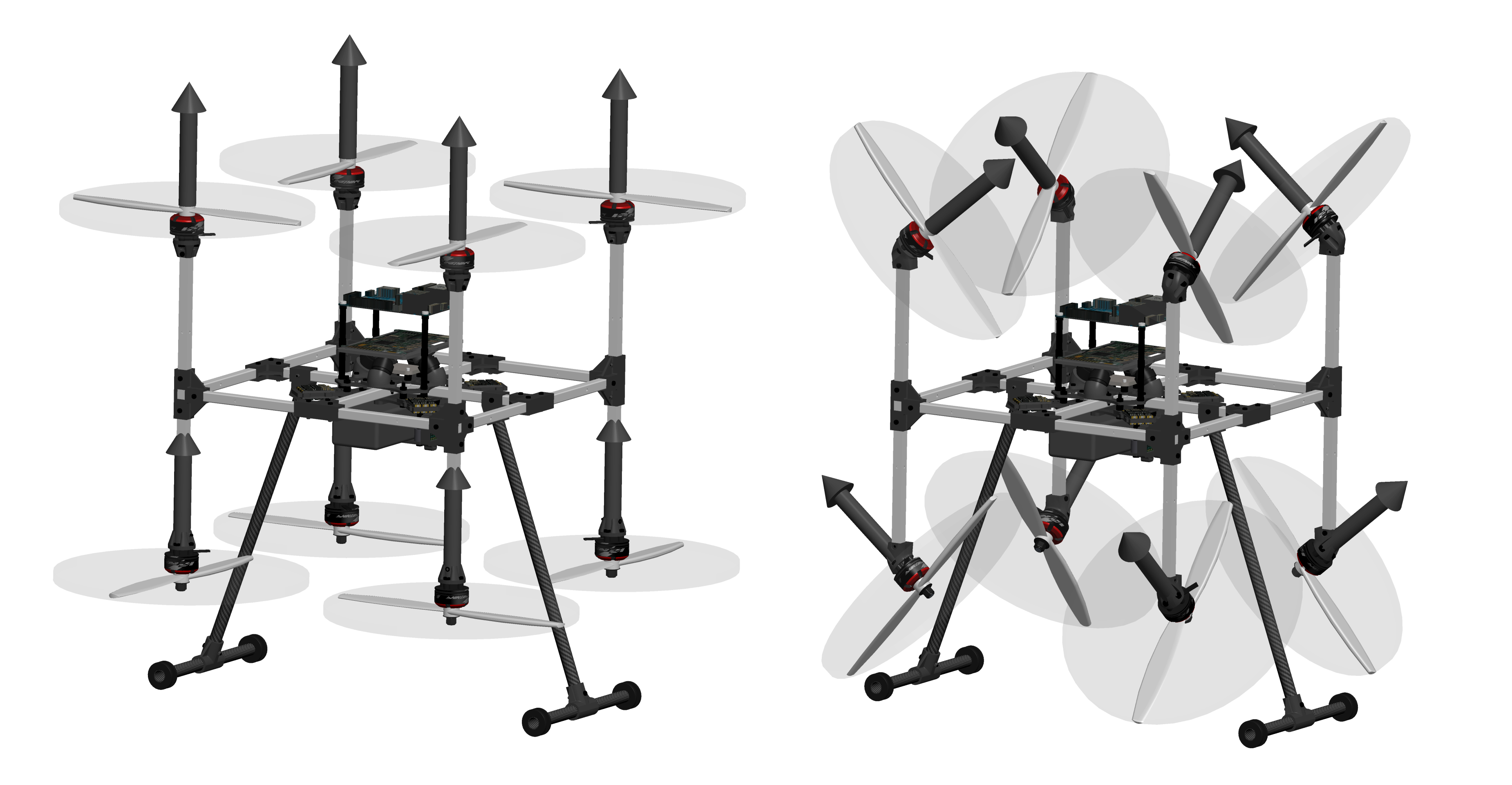}
    \caption{OmniMorph:  \acrshort{ua} configuration (left) and \acrshort{od} configuration (right).}
\label{fig:morphing}
\end{figure}
		This led to the development of new \glspl{mrav} and especially to the exploration of designs different from the popular underactuated multirotors that use fixed coplanar/collinear propellers (i.e.,  quadrotors, hexarotors, and their variants). The new designs rely on choosing the number, position, and orientation of the propellers to provide the multirotor with different actuation properties and abilities to achieve a specific task \cite{hamandi_design_2021}. \textcolor{black}{Morphing designs with varying arm length and consequent proper controller tuning have also been proposed in the literature \cite{kose2023simultaneous, csahin2022simultaneous,kose2020simultaneous,kose2023simultaneous2}.}
%

%
Recently, several works addressed the design of \textit{omnidirectional} multirotors. Omnidirectionality is the property of the robot to sustain its weight in any orientation. In the literature, different actuation strategies to achieve omnidirectionality have been proposed. In \cite{hamandi_understanding_2021}, a novel metric to assess the omnidirectional property of a generic multi-rotor aerial vehicle was presented. This metric is calculated numerically from the force set
of the platform, and as such it relies on the platform geometry and thrusters; moreover, it allows a direct assessment of a platform’s omnidirectional property given its weight. \cite{hamandi_understanding_2021} also showed the use of this metric in the process of upgrading the design presented in \cite{hamandi_omni-plus-seven_2020}, by placing the propellers on a virtual sphere centered around the platform’s 
\textcolor{black}{Center of Mass (\acrshort{com})}.
In the following, we categorize omnidirectional designs into three main classes. 
	
	The first class uses fixedly tilted \emph{bi}-directional propellers (which generate thrust in both directions by inverting their sense of rotation). The authors in \cite{brescianini_design_2016} presented one of the first omnidirectional prototypes, where they placed eight bi-directional propellers on the edges of a cube. The novelty of their design is the optimization-based placement and orientation of the propellers, where the cube shape was chosen to ensure a rotation-invariant inertia tensor. Another omnidirectional design with fixed propellers was presented in \cite{park_design2016} and \cite{park_odar2018}, achieving omnidirectional wrench generation with six and eight bi-directional propellers, respectively.

	The second class, like the design proposed in~\cite{tognon_omnidirectional2018}, 
	optimizes the orientation of fixed \emph{uni}-directional propellers (i.e., able to produce thrust in only one direction, or, in other words, to rotate in only one sense) to achieve omnidirectionality while equally distributing the thrust produced by each propeller and aiming for an isotropic (sphere-like) shape of the corresponding force set. In fact,  \cite{tognon_omnidirectional2018} showed that the minimal number of 7 (seven) uni-directional propellers is necessary to achieve omnidirectional position and orientation tracking. Later,  \cite{hamandi_omni-plus-seven_2020} presented a working prototype of the design with experiments.
	
The third class of omnidirectional platforms relies on a morphing design. Uni- or bi-directional propellers  \emph{actively tilted} by servo motors are used. \cite{ryll_novel2015} proposes a quadrotor with propellers individually tilting about their radial axis. The authors of \cite{kamel_voliro2018} use the same idea (independently controlled radially tilting thrusters) for an hexarotor. Another version of their design was presented in \cite{allenspach_design_2020}, in which double propeller groups were used, two propellers rotating in opposite directions on each rotation axis to increase the platform efficiency and payload capacity in addition to increased force bounds along different directions. 

\textcolor{black}{Design aimed at optimizing control effort is an established approach in aerial platforms \cite{sal2022simultaneous} and it is even more critical for multirotor aerial robots, as they are typically characterized by limited flight endurance.} The advantage of the third design approach relies on the possibility of optimizing the energy consumed for a given desired orientation, which is obtained by aligning together as much as possible the propeller spinning directions for a given orientation, in a way that each propeller tends to be as much counter aligned as possible to the gravity force. This also allows for balancing dexterity and energy efficiency based on the needs imposed by specific tasks \cite{ryll_modeling_2016} \cite{ryll_fast-hexmorphing_2022}.  Indeed, the configuration in which all the directions are aligned is more energetically efficient. In that configuration, the propeller forces are all parallel to each other but the thrust is uni-directional; when this is not the case, there are propeller forces that cancel each other without actively contributing to the motion. That is the price paid to achieve multi/omnidirectionality.

From a control allocation point of view, fixed-propeller multirotors in \cite{brescianini_design_2016},  \cite{park_design2016}, \cite{hamandi_omni-plus-seven_2020} are controlled by (pseudo)inverting the wrench map between the propellers' speeds and the total body wrench. Those systems are affine in the control inputs and the tilting angles of the propellers are constant design parameters. A similar control approach is used in \cite{park_odar2018}, where a different rotor input allocation is proposed, based on the infinity-norm minimization; in this way, propellers' speed constraints are implicitly taken into account, even though with no guarantees. 

For the second type of omnidirectional platforms, in \cite{hamandi_omni-plus-seven_2020} a dynamic inversion plus input allocation strategy is proposed. The input allocation modifies the minimum norm solution found by inverting the allocation matrix to ensure that the rotational speeds are positive. \textcolor{black}{For a fixed-propeller hexarotor that is fully actuated but not omnidirectional, \cite{bicego2020nonlinear} proposes a Nonlinear Model-Predictive Controller.}

The control of the third class of platforms is challenging, as the tilting angle must also be determined, which appears nonlinearly in the dynamics. Feedback linearization at the jerk level has been proposed \cite{ryll_novel2015}, thanks to the fact that the time-differentiation the system is affine in the (new) control inputs. In \cite{kamel_voliro2018}, a change of variables is used to obtain a static allocation matrix not dependent on the tilting angle; that is pseudo-inverted, and the initial inputs are then computed. Both these methods require the full allocation matrix to be non-singular and may result in unfeasible inputs due to neglected input bounds. \cite{allenspach_design_2020} proposes a jerk-level control in which the optimal control inputs are obtained by weighted pseudo-inversion of the full allocation matrix. As input constraints are not considered, the inputs are saturated before being integrated and sent to the actuators. The control method avoids singularities by adding a bias to the desired tilting angle in case of a large condition number of the wrench map matrix. 


In this work, we overcome some of the limitations of existing concepts and controllers of omnidirectional platforms. Specifically, the contributions of this work are as follows.
\begin{itemize}
    \item A novel omnidirectional morphing platform design concept, OmniMorph, is proposed. The platform belongs to the third class described in this section and exploits the synchronized drive of all propellers by only one servo motor. This choice allows keeping at the bare minimum the additional payload used to carry additional servo motors in the other designs seen before.
    \item The actuation properties of  OmniMorph are formally studied. 
    \item A novel control method is proposed for multi-rotors with actively and synchronously tilting propellers. Compared to the state of the art, the proposed controller allows automatic transitions between underactuated and omnidirectional configurations while explicitly accounting for input constraints, only indirectly handled through redundancy resolution in the literature \cite{ryll_novel2015}. 
    \item The proposed design and controller are tested in a realistic simulation environment.
    \item \textcolor{black}{A preliminary prototype is built and tested to assess the mutual aerodynamic interference among the propellers. Hence, simulations under degraded and unknown propeller performance are carried out to assess the robustness of the proposed controller.}
\end{itemize}
\section{OmniMorph Design and model}\label{Sec:Des&Model}
\subsection{Concept and Design}
The OmniMorph platform consists of 8 bi-directional propellers attached to the main body frame, placed on the vertices of a cube centered at the \acrshort{com}, as in Figure \ref{fig:morphing}. The idea of this work is to design a morphing platform, i.e., capable of tilting its propellers, that can span while flying, a variety of configurations in between two extreme cases: a dexterous but less energetically efficient omnidirectional configuration and an efficient but underactuated unilateral thrust configuration. More than that, we ask that all propellers tilt in a synchronized way, meaning they can be all attached to a single extra actuation unit (a tilting mechanism) driven by only one servo motor. That reduces the complexity and the weight of the prototype, allowing for a higher payload. 

\begin{figure*}[t]
    \centering
\includegraphics[width=\textwidth]{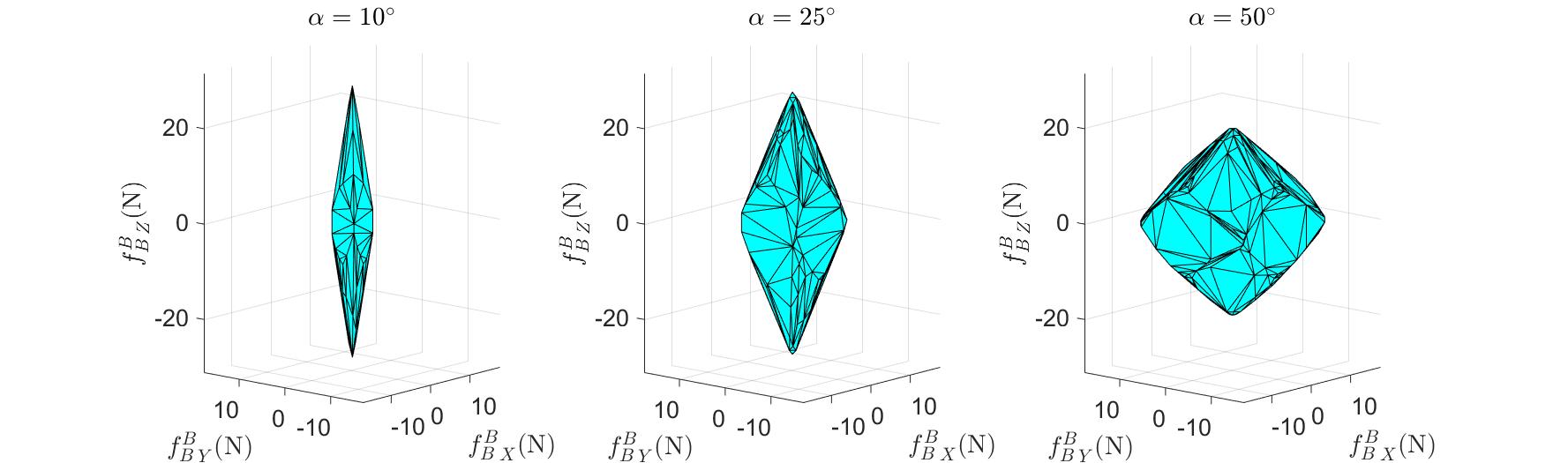}
    \caption{Sets of feasible forces for different values of $\alpha$. The three components of the thrust in the body frame are displayed. OmniMorph is omnidirectional when a sphere with a radius equal to the robot's weight is inscribed inside the polytope of feasible forces.}
    \label{fig:feasible_forces}
\end{figure*}

Note that each propeller tilts about a different axis of the same amount, which we call the \emph{tilting angle} $\alpha.$  The change in the orientation of the propellers (morphing), changes (i.e., it morphs) the force and moment feasible sets. When all rotors are parallel to each other (i.e. titling angle $\alpha$ equal to zero) the force set is a thin line, i.e., the generated total thrust can be produced only along one axis,  
making the vehicle underactuated (\acrshort{ua})\textemdash see Figure~\ref{fig:morphing} left. In this case, the platform is, from the actuation point of view, equivalent to the standard underactuated quadrotor, hexarotor, and so on: only 4 \textcolor{black}{Degrees of Freedom (\acrshort{dof})} are independently controlled, the 3D position and the heading angle. This configuration is the most energetically efficient and it is recommended in tasks in which the control of the full orientation or full wrench is not essential.

When the tilting angle $\alpha$ assumes positive values, the feasible force set morphs into a polyhedron with non-zero volume. The larger the tilting angle the larger the volume. The total thrust can be generated in all three orthogonal directions, to a certain extent, depending on the tilting angle value. Ultimately, the vehicle can assume an \acrshort{od} configuration similar to the vehicle developed in \cite{brescianini_design_2016}\textemdash see Figure~\ref{fig:morphing}, right. Such a  configuration allows controlling the 6D full pose of the vehicle sustaining its weight in any possible orientation.  A representation of the polytopes of feasible forces of OmniMorph for different values of $\alpha$ computed as in \cite{hamandi_design_2021} is given in Figure \ref{fig:feasible_forces}.

\begin{figure}[t]
    \centering
  \includegraphics[width=0.8\columnwidth]{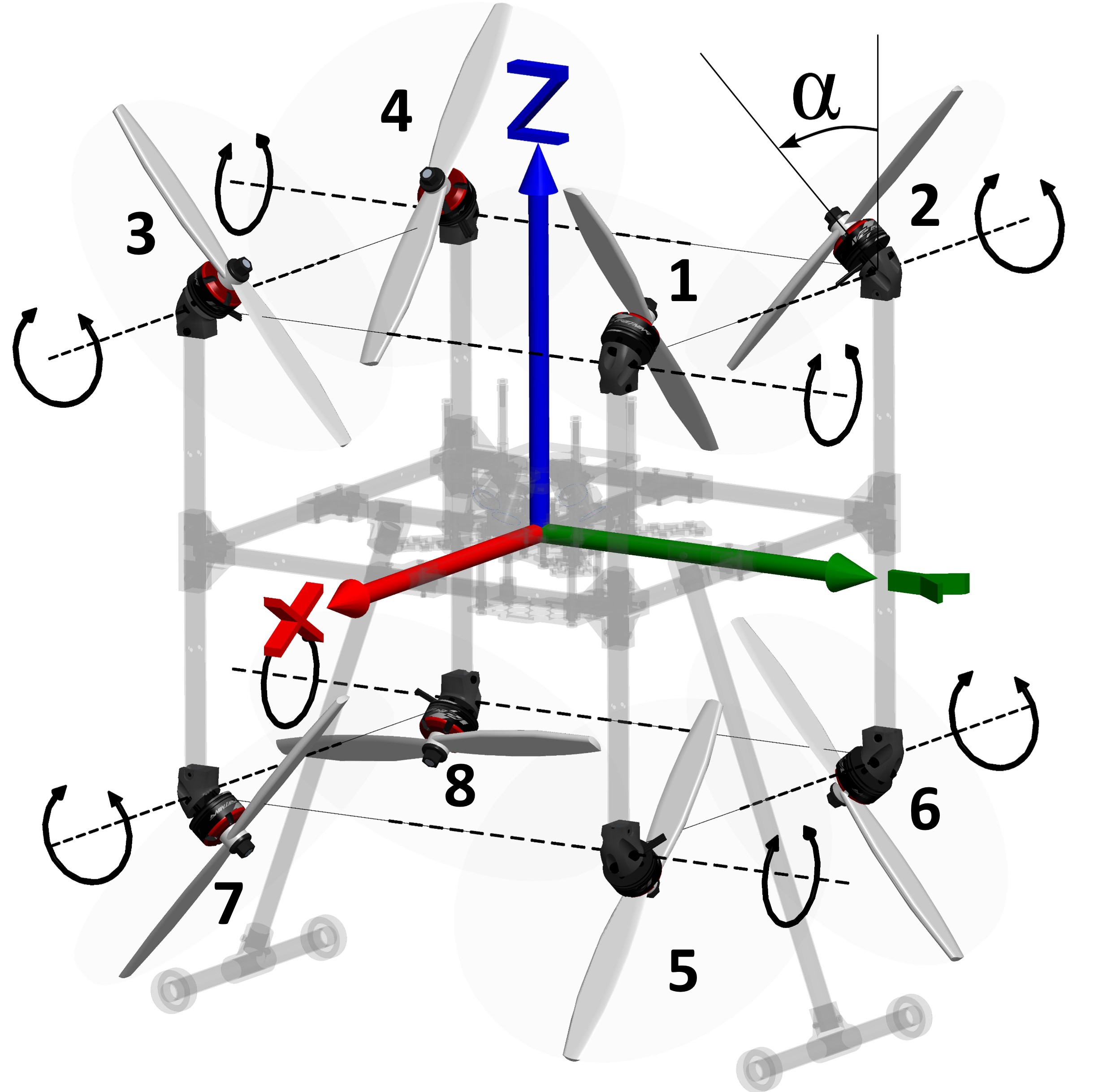}
    \caption{Representation of OmniMorph in which the axes of rotation of the propellers, lying along the cube edges, are highlighted. }
    \label{fig:omni-rot}
\end{figure}

In the following, the design concept of OmniMorph is detailed. 
We define a reference frame attached to the robot as a set of an origin and three orthogonal axes: ${\mathcal{F}_{B}^{}: \{ O_B, \mathbf{x}_{B}^{}, \mathbf{y}_{B}^{}, \mathbf{z}_{B}^{} \}}$; \textcolor{black}{the origin, $O_B$, is coincident with the robot \acrshort{com} and the three axes, $\mathbf{x}_{B}^{},$ $\mathbf{y}_{B}^{},$ and $\mathbf{z}_{B}^{}$, are aligned with the edges of a cube of side length $L$. Let us define a point at the $i$-th propeller's center as $O_{p_{i}}$, and its position in the body-fixed frame as $\vect{p}_{p_{i}}^{B}$. Then, we define, as done in \cite{brescianini_omni-directional_2018}, the matrix $\matr{P}=[\vect{p}_{p_{1}}^{B},...,\vect{p}_{p_{8}}^{B}]$. Particularly, we have that } \begin{equation}\matr{P}_p=\frac{L}{\sqrt{3}}\begin{bmatrix}  1&-1&1&-1&1&-1&1&-1\\1&1&-1&-1&1&1&-1&-1\\1&1&1&1&-1&-1&-1&-1
\end{bmatrix},\label{eq:prop_pos}\end{equation} where the propellers are numbered as in Figure \ref{fig:omni-rot}.

\textcolor{black}{In \cite{brescianini_omni-directional_2018}, an optimal \textit{fixed} orientation of the propellers to attain omnidirectionality is found. Especially, in \cite{brescianini_omni-directional_2018}, the authors define the coordinates of the spinning axis of their \textit{fixedly tilted} propellers, let us call them ${\tilde{\vect{z}}_{i}}^{B}$, 
and they collect them into the matrix $\matr{B}=[\tilde{{\vect{z}}_{1}}^B \cdots {\tilde{\vect{z}}_{8}}^B]$} given by \begin{equation}\matr{B}=\begin{bmatrix}-a&b&-b&a&a&-b&b&-a\\b&a&-a&-b&-b&-a&a&b\\c&-c&-c&c&c&-c&-
c&c\end{bmatrix},\label{eq:Bresc_att}\end{equation}
where $a=\frac{1}{2}+\frac{1}{\sqrt{12}},$ $b=\frac{1}{2}-\frac{1}{\sqrt{12}},$ and $c=\frac{1}{\sqrt{3}}$.

\textcolor{black}{Let us now define as ${\tilde{\bm{b}}_{i}}$ the coordinates of the unit vector, centered on $O_{i}$, that describes the axis of rotation of propeller $i-$th to bring its spinning axis from the direction described in \eqref{eq:Bresc_att} 
to a configuration in which it is parallel to ${\tilde{\mathbf{z}}_{i}}$, thus bringing the robot from the configuration described in \cite{brescianini_omni-directional_2018} into the unilateral-thrust configuration.} 
We have that $\tilde{\bm{b}}_{i}$ is given by the cross
product between $\tilde{\vect{z}}_{i}$ and $\vE{3}$, 
where $\vE{i}$ is the $i-$th column of the identity matrix $\eye{3}\in\mathbb{R}^{3\times3}.$ 
Very interestingly, $\tilde{\bm{b}}_{i}$ is orthogonal to $\tilde{\bm{b}}_{j}$ $\forall i\in\{1,4,5,8\}$, $\forall j\in\{2,3,6,7\},$ \textcolor{black}{similarly to Figure \ref{fig:omni-rot}, where the dotted lines representing the rotational axes of the propellers are orthogonal to each other.} More than that, \textcolor{black}{ we note that by rotating the cube \textcolor{black}{generated by} those axes }of $\frac{\pi}{12}$ radians (i.e., $15\,\rm{deg}$) around $\mathbf{z}_{B}^{}$, one obtains that the propellers' tilting axes lie along the edged of the cube \textcolor{black}{formed by the propellers. Let us call these new propeller tilting axes $\bm{b}_{i}$.} Such a symmetric configuration greatly simplifies the mechanical realization of the tilting mechanism, and, as it will be more clear in the following, \textcolor{black}{it allows for transitioning between \acrshort{udt} and omnidirectional configurations. Eventually, that configuration is conveniently chosen for OmniMorph and is reported in Figure \ref{fig:omni-rot}}. 


%

To summarize formally, the positions of the centers of the propellers are given in \eqref{eq:prop_pos}, and the attitude is described as follows. Define a frame attached to each propeller ${\mathcal{F'}_{i}^{} : \{ O' _{i}, \mathbf{x'}_{i}^{}, \mathbf{y'}_{i}^{}, \mathbf{z'}_{i}^{} \}}$, where $\mathbf{x'}_{i}^{}=\tilde{\bm{b}}_{i}$, $\mathbf{z'}_{i}^{}=\mathbf{z'}_{B}^{},$ and $\mathbf{y'}_{i}^{}$ is chosen  orthogonal to the first two. Denote the rotation matrix between ${\mathcal{F}_{B}^{}}$ and ${\mathcal{F'}_{i}^{}}$ with $\mathbf{R'}_{i}$. 
Eventually, the rotation between ${\mathcal{F}_{B}^{}}$ and each propeller's  frame \textcolor{black}{of OmniMorph,} ${\mathcal{F}_{p_{i}}^{} : \{ O_{p_{i}}, \mathbf{x}_{i}^{}, \mathbf{y}_{i}^{}, \mathbf{z}_{i}^{} \}}$ is given by ${^B\mathbf{R}_{i}(\alpha)=\mathbf{R'}_{i}\mathbf{R}_{Z}\left(\frac{\pi}{12}\right)\mathbf{R}_{X}(\alpha),}$ where $\mathbf{R}_{X}()$ $\mathbf{R}_{Z}()$ are the elementary rotations about the $X$  and $Z$ axes. Each propeller exerts its thrust along the axis $\mathbf{z}_{i}^{}$ which is rotated by $\alpha$ about an axis lying along the edges of the cube that connects the propellers' centers.

\begin{table*}[t]
    \centering
    \begin{tabular}{ |c|c|c| } \includegraphics[trim={2cm 9cm 2cm 9cm},clip, width=0.60\columnwidth]{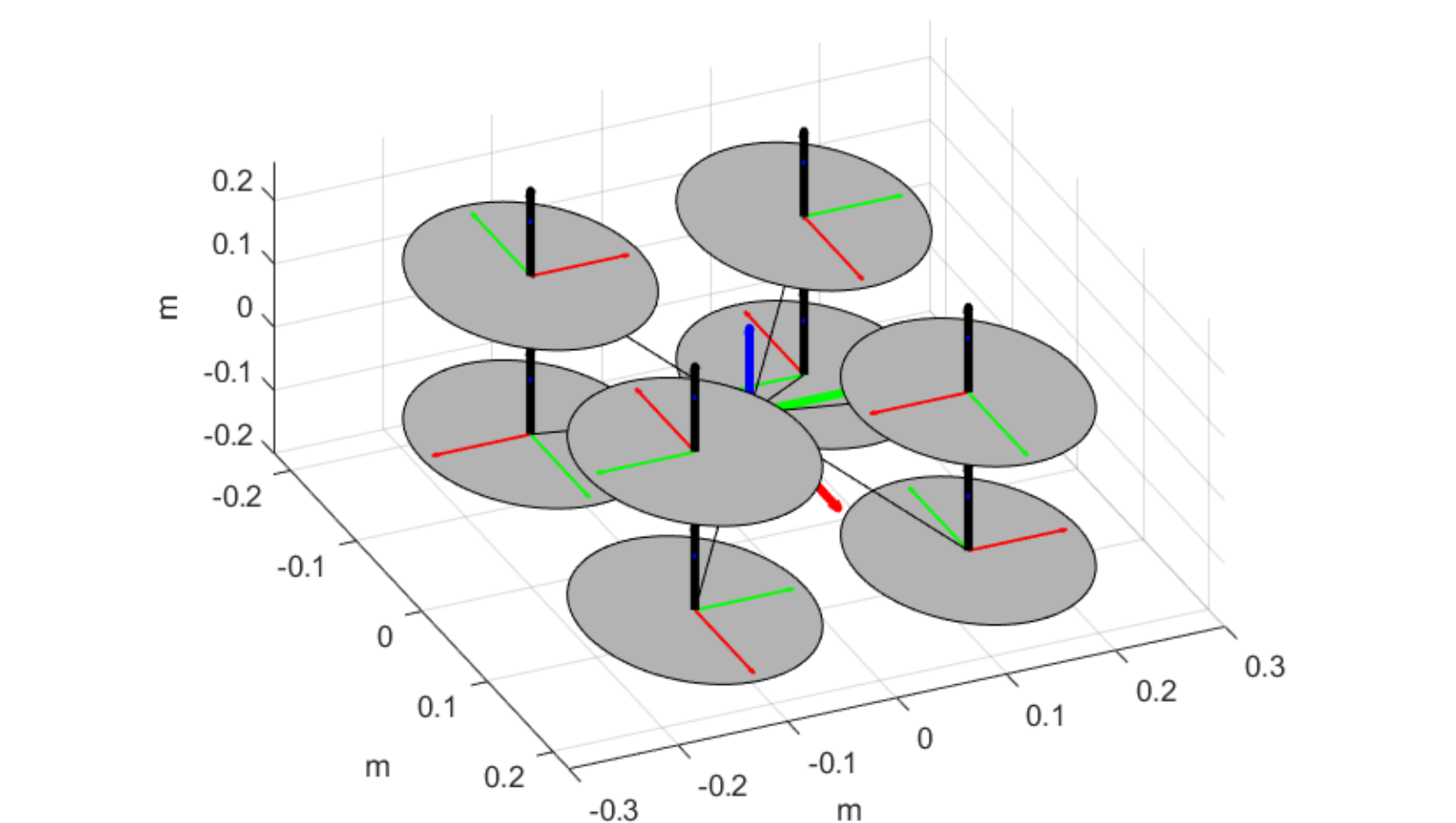}&
\includegraphics[trim={2cm 9cm 2cm 9cm},clip, width=0.60\columnwidth]{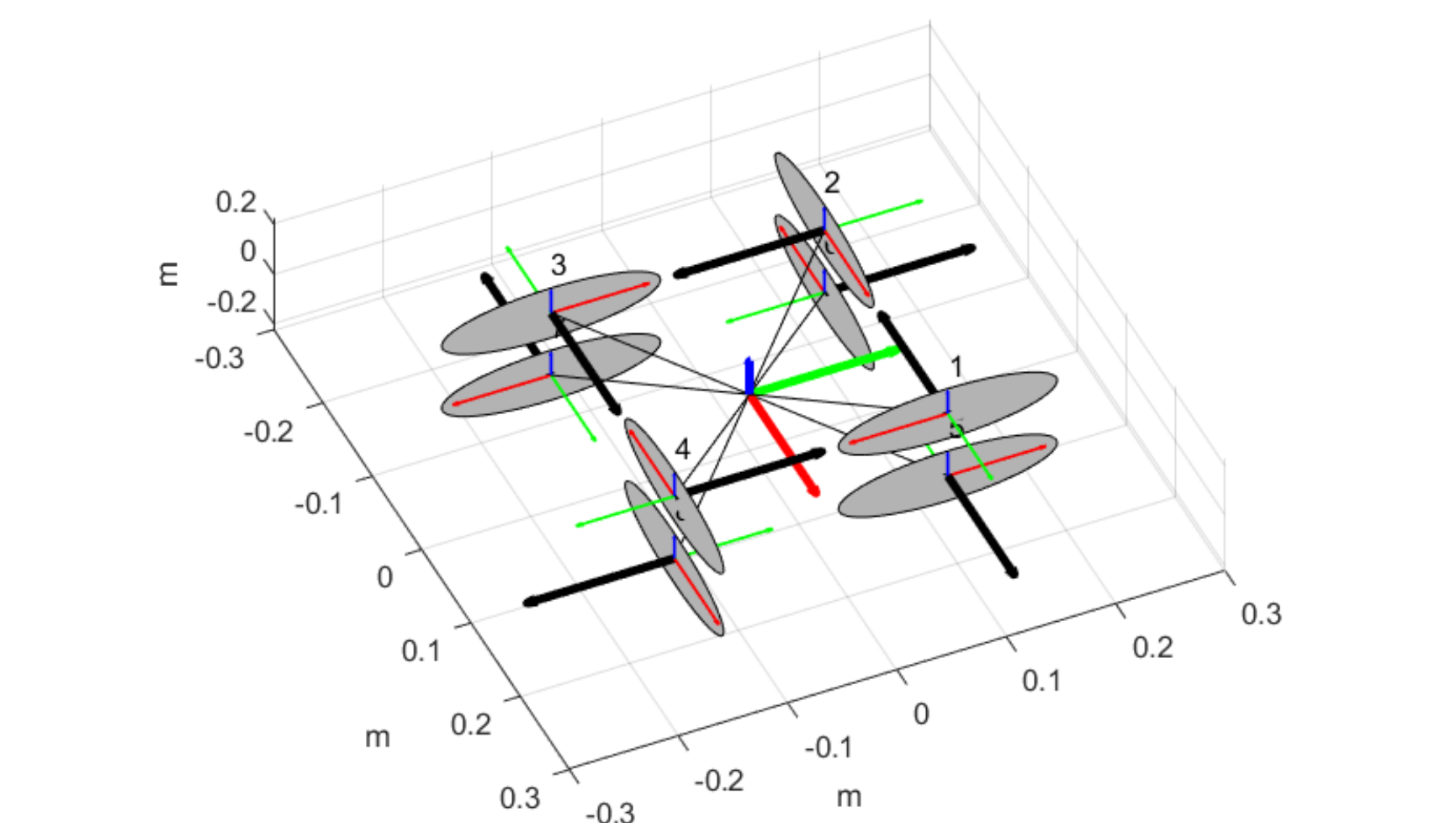} &  \includegraphics[trim={2cm 9cm 2cm 9cm},clip, width=0.60\columnwidth]{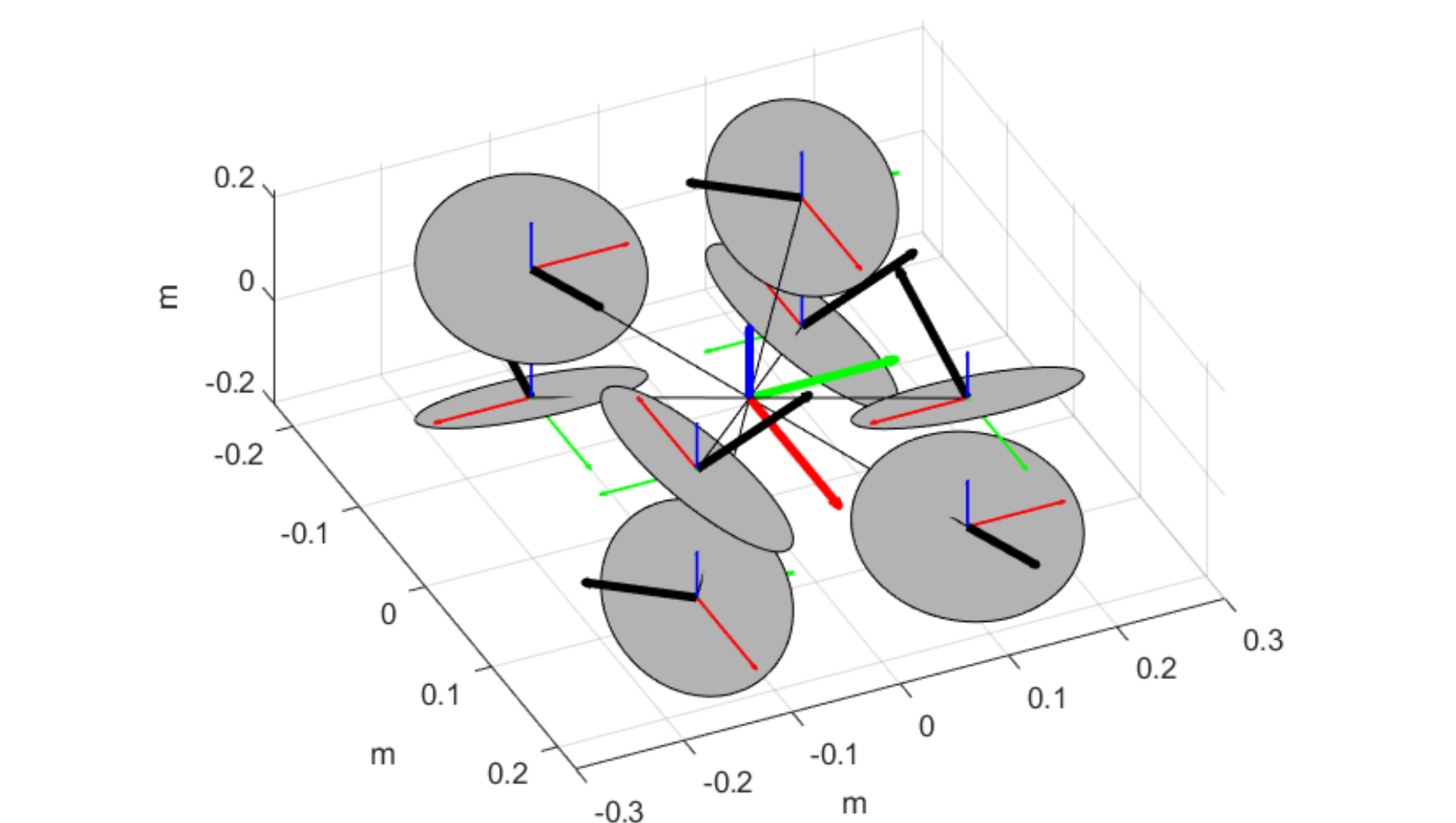}\\
\hline $\alpha=0$ & $\alpha=\frac{\pi}{2}$ &  $\alpha\neq\frac{\pi}{2},\pi,0$ \\ \hline Underactuated & Fully Actuated unless $\bm{u}_\omega=0$ &  Fully Actuated, Redundant
\end{tabular}
    \caption{Actuation properties for different values of $\alpha$. Grey circles are the propellers' planes. Each is associated with a local frame, and a body-fixed frame is in the middle of the robot. The RGB convention is used to indicate the x-, y-, and z-axis, respectively. The direction of the thrust produced by the individual propellers is represented with a thick black arrow.}
    \label{tab:actuation}
\end{table*}
\subsection{Model of the OmniMorph} In this section, the dynamics equations of OmniMorph are given. 
Let us define an inertial frame  $\mathcal{F}_{W}^{} : \{ O_{W}^{}, \mathbf{x}_{W}^{}, \mathbf{y}_{W}^{}, \mathbf{z}_{W}^{} \}$. 
 The position of $O_{B}^{}$ in $\mathcal{F}_{W}^{}$ is indicated as $\vect{p}$ and its attitude is expressed compactly as  $\mathbf{R}:=^W\mathbf{R}_B$.\footnote{In general, we indicate as $^A\mathbf{R}_{A'}$ the rotation from frame $\mathcal{F}_{A}^{}$ to  $\mathcal{F}_{A'}^{}$.}

 The dynamic equations of the robot, expressed as Newton-Euler equations of an actuated rigid body, are as follows, where the translational dynamics is expressed in the inertial frame and the rotational dynamics in the body frame. We indicate as $\boldsymbol{\omega}$ the angular velocity of $\mathcal{F}_{B}^{} $ w.r.t. $\mathcal{F}_{W}^{}$  expressed in $\mathcal{F}_{B}^{}$, with $m$ the robot mass, and with $\mathbf{J}\in\mathbb{R}^{3\times3}$ its rotational inertia.

\begin{equation}
\begin{bmatrix}
m \mathbf{I}_{3}^{} & \boldsymbol{0}_{3}^{} \\
\boldsymbol{0}_{3}^{} & \mathbf{J}
\end{bmatrix}
\begin{bmatrix}
\ddot{\vect{p}} \\
\dot{\boldsymbol{\omega}}
\end{bmatrix}
= 
\begin{bmatrix}
-m g \vect{e}_{3}^{} \\
-\boldsymbol{\omega} \times    \mathbf{J} \boldsymbol{\omega}	
\end{bmatrix} +
\begin{bmatrix}
\mathbf{R} & \boldsymbol{0}_{3}^{} \\
\boldsymbol{0}_{3}^{} & \mathbf{I}_{3}^{}
\end{bmatrix}
\begin{bmatrix}
\vect{f}_{B}^{B}  \\
\boldsymbol{\tau}_{B}^{B} 
\end{bmatrix}
\label{eq:N-E}
\end{equation}
 $\vect{f}_{B}^{B}$  and $\boldsymbol{\tau}_{B}^{B}$ are the total force and torque applied to the \acrshort{com}, expressed in 
${\mathcal{F}_{B}^{}}$.  

The rotational kinematics of the robot is: 
\begin{equation}
\dot{\mathbf{R}} = \mathbf{R} [\boldsymbol{\omega}]_{\times}. 
\end{equation}
where $[\boldsymbol{\omega}]_{\times}$ is the skew symmetric matrix associated to $\boldsymbol{\omega}$.
Each rotor produces a thrust force and a drag moment which we model, similarly to \cite{michieletto2018fundamental}, as follows.
\begin{equation}\label{eq:thrust_drag}
\vect{f}_{i}^{} =  c_{f}^{}w_{i}|w_{i}|\mathbf{z}_{i}^{}, \quad  \bm{\tau}_{d_{i}} = -k_i c_{\tau}w_{i}|w_{i}|\mathbf{z}_{i}^{}, 
\end{equation}
where $c_f,c_{\tau}>0$ are the thrust and drag coefficient, respectively, and $w_i$ is a scalar with a module equal to the norm of the propeller angular velocity and a sign defined such that $w_i>0$ when the thrust produced by the propeller has the same direction of $\mathbf{z}_{i}^{}$. The drag moment is always opposite to the angular velocity of the propeller; as a consequence,  $k_i=-1$ for propellers with descending chord (i.e., producing thrust in the same direction of the angular velocity, also called `counterclockwise')  and $k_i=+1$ for
propellers with ascending chord (i.e., producing thrust in the opposite direction of the angular velocity, also called `clockwise').
 The applied body forces and torques are the results of all propellers and their orientation:
\begin{equation}\label{eq:f_iB}
\vect{f}_{B}^{B} = \sum_{i=1}^{n}	\vect{f}_{i}^{B} =  \sum_{i=1}^{n}  \prescript{B}{}{\mathbf{R}_{p_{i}}^{}}  \vect{e}_{3} f_{i}^{} 
\end{equation}
%
\begin{equation}\label{eq:tau_iB}
\begin{split}		
\boldsymbol{\tau}_{B}^{B} & = \sum_{i=1}^{n}	\boldsymbol{\tau}_{f_{i}}^{B} + \boldsymbol{\tau}_{d_{i}}^{B}  =  \sum_{i=1}^{n}  \vect{p}_{p_{i}}^{B} \times \vect{f}_{i}^{B} - k_{i}c_{f}^{\tau} \vect{f}_{i}^{B}  \\
& = \sum_{i=1}^{n} ( [\vect{p}_{p_{i}}^{B}]_{\times} - k_{i}c_{f}^{\tau} \mathbf{I}_{3}) \prescript{B}{}{\mathbf{R}_{p_{i}}^{}} \vect{e}_{3}^{} f_{i}^{}
\end{split}
\end{equation}
where $c_{f}^{\tau}=c_{\tau}/c_{f}$.
We indicated with 	$\vect{f}_{i}^{B}$ the thrust force of the $i$-th propeller expressed in $\mathcal{F}_{B}^{}$, and with $\boldsymbol{\tau}_{f_{i}}^{B}$ and  $\boldsymbol{\tau}_{d_{i}}^{B}$
the torque produced by the same propeller's thrust at the \acrshort{com} and the effect of the propeller's drag torque at the \acrshort{com}, respectively.


We denote with $\boldsymbol{u}$ the input of the system, consisting of the single morphing angle ${\alpha}$ and the rotor speeds contained in the vector $\boldsymbol{u}_{w}$ such that 
\begin{equation}
  \boldsymbol{u}_{w} = [u_{w_{1}}  \cdots u_{w_{n}} ]^\top= [w_1|w_1|  \cdots w_n|w_n| ]^\top \in \mathbb{R}^n.
\end{equation}
Hence, one has
$
\boldsymbol{u} = \begin{bmatrix}
\boldsymbol{u}_{w}^\top& \alpha   
\end{bmatrix}^\top\in \mathbb{R}^{9}.
$
Based on \eqref{eq:f_iB} and \eqref{eq:tau_iB}, we have that the actuation wrench is
\begin{equation}\label{eq:wrench_map}
\mathbf{w}(\boldsymbol{u})=\begin{bmatrix}
\vect{f}_{B}^{B}  \\
\boldsymbol{\tau}_{B}^{B} 
\end{bmatrix} = 
 \mathbf{A}(\alpha)  \boldsymbol{u}_{w}.  
\end{equation}
The full expression of  $\mathbf{A}(\alpha)$ is at the end of Sec. \ref{sec:act_prop}.

%
\subsection{Actuation Properties}\label{sec:act_prop}
First, we define the \textit{full allocation matrix}  ${\mathbf{F}(\boldsymbol{u}) \in \mathbb{R}^{6\times9}}$  as in  \cite{hamandi_design_2021}: 
\begin{equation}\label{eq:full_all_short}\mathbf{F}(\boldsymbol{u})=
\begin{bmatrix} 
\mathbf{F}_1(\alpha) & \mathbf{F}_2(\boldsymbol{u}_{w}, \alpha)\end{bmatrix}
\end{equation}
%
with  $
\mathbf{F}_1(\alpha) = \frac{\partial\mathbf{w}}{\partial\boldsymbol{u}_{w}}(\alpha) \in \mathbb{R}^{6\times 8}, \mathbf{F}_2(\boldsymbol{u}) = \frac{\partial\mathbf{w}}{\partial{\alpha}}(\boldsymbol{u}) \in \mathbb{R}^{6}.$
\textcolor{black}{Note that $\mathbf{F}_1$ only depends on $\alpha$ because $\mathbf{w}$ in \eqref{eq:wrench_map} depends \textit{linearly} on $\vect{u}_w$. Particularly, following the definition given above and considering \eqref{eq:wrench_map}, we have that $\mathbf{F}_1=\mathbf{A}(\alpha)$ and $\mathbf{F}_2=\frac{\partial(\mathbf{A}(\alpha)\boldsymbol{u}_w)}{\partial\alpha}.$} 
Following the method provided in \cite{hamandi_design_2021}, we study the actuation properties of the platform by looking at the rank of  $\mathbf{F}(\boldsymbol{u})$ in \eqref{eq:full_all_short}. For all the points in the input space in which $\rank(\mathbf{F}(\boldsymbol{u}))=6$ (full rank), the platform is fully actuated. In those points, the actuation wrench can be changed in every direction by suitably acting on the input. Instead, in all points in which  $\rank(\mathbf{F}(\boldsymbol{u}))<6$ (rank deficient) the platform is underactuated, this means that when the input is in those singular points there are some forbidden directions along which the actuation wrench cannot be changed. 

First, we note that studying the rank of \textcolor{black}{$\mathbf{F}_1(\alpha)$ helps us study the rank of the full allocation matrix, $\mathbf{F}(\boldsymbol{u})$}: 
\begin{itemize}
    \item $\rm{rank}(\mathbf{F}_1(\alpha))=4$ is a sufficient condition for the platform to be underactuated: the rank of the full allocation matrix $\mathbf{F}(\bm{u})$ can be at most $5<6$.
    \item $\rm{rank}(\mathbf{F}_1(\alpha))=6$ is a sufficient condition for the system to be fully actuated;
    \item If $\rm{rank}(\mathbf{F}_1(\alpha))=5$, the full-/under-actuation needs further analysis of $\mathbf{F}_2(\boldsymbol{u})\in\mathbb{R}^{6\times1}$ in relation to $\mathbf{F}_1(\alpha)$.
\end{itemize}

Hence, we study the rank of $\mathbf{F}_1(\alpha)$ for various values of $\alpha$, finding that
\begin{equation}\rm{rank}(\mathbf{F}_1(\boldsymbol{\alpha}))=\begin{cases} 4 & \rm{if}\ \alpha=0+k\pi, k\in\mathbb{N} \\ 5 & \rm{if} \ {\alpha}=\frac{\pi}{2}\\ 6 & \  \rm{otherwise}\\\end{cases}\end{equation}

So, from what has been said so far,  OmniMorph is underactuated for $\alpha=0$.   
Moreover, ${\rm{rank}} ({\mathbf{F}}{(\bm{u}_w,0))}=5$ if $\boldsymbol{u}_{w}\neq\bm{0}$ and if not all ${u_{w}}_i$ are equal to each other, i.e., the total wrench can be instantaneously changed on a `reduced' five-dimensional manifold.

For $\alpha=\pi/2$, OmniMorph is in general under-actuated. However, it is  fully actuated ($\rm{rank} (\mathbf{F}(\boldsymbol{u}_{w}, \pi/2))=6)$ if  $\boldsymbol{u}_{w}\neq\bm{0}$. An intuitive interpretation is provided in the following. When $\alpha=\pi/2,$ all propellers' thrusts are on the same plane in $\mathcal{F}_{B}^{}$. 
%

The yaw acceleration can be changed as the individual thrusts generate a torque around the vertical axis of the robot. The pitch and roll accelerations can be changed thanks to the drag moment of the propellers. The translational acceleration in the horizontal plane is clearly changeable by the thrust, but the translational acceleration along $\mathbf{z}_{B}^{}$ is not. As soon as $\alpha$ changes, the propellers' thrusts have a component in the vertical direction as well, which gives control of the vertical acceleration, unless the propellers' thrusts are actually zero (mathematically, $\boldsymbol{u}_{w}=\bm{0}$). 

Even if the platform is fully actuated it might not be able to fly properly because the propeller thrust intensity may be not large enough to sustain the platform's weight. 

Eventually, for all other values of $\alpha$, the platform is fully actuated. A schematic summary of the OmniMorph actuation properties is in Table~\ref{tab:actuation}. 

For sufficiently large values of $\alpha,$ the polygon of the achievable forces, see Fig.~\ref{fig:feasible_forces} contains a sphere whose radius corresponds to the weight of the platform: the robot can sustain its weight in any orientation;  thus, it is omnidirectional. 

\textcolor{black}{For the sake of completeness, the full allocation matrix is reported in the following, where $s_\alpha$ and $c_\alpha$ indicate the sin and cosine of $\alpha$, respectively}:
 \begin{align*}&\mathbf{F}={c_f}\left[\scalebox{0.90}[1]{$\begin{matrix}
    -s_\alpha& 0 &  s_\alpha & 0\\ 
   0&s_\alpha &-s_\alpha&0\\     
   c_\alpha&   c_\alpha& c_\alpha& c_\alpha\\
   Lc_\alpha - \frac{c_\tau s_\alpha}{c_f} & Lc_\alpha + Ls_\alpha& \frac{c_\tau s_\alpha}{c_f} - Lc_\alpha&       - Lc_\alpha - Ls_\alpha\\    
   - Lc_\alpha - Ls_\alpha & L  c_\alpha + \frac{ c_\tau  s_\alpha}{c_f} &       L  c_\alpha + L  s_\alpha & - L  c_\alpha - \frac{ c_\tau  s_\alpha}{ c_f} \\    L   s_\alpha +  \frac{ c_\tau   c_\alpha}{c_f} & L   s_\alpha -  \frac{ c_\tau   c_\alpha}{c_f} & L   s_\alpha +  \frac{ c_\tau   c_\alpha}{c_f} &   L   s_\alpha -  \frac{ c_\tau   c_\alpha}{c_f} \\  
    \end{matrix}$}\right. \\
    &\left.\scalebox{0.85}[1]{$\begin{matrix}s_\alpha & 0 & 0 & -s_\alpha\\ 0& -s_\alpha&s_\alpha&0&\\   c_\alpha&                          c_\alpha&                          c_\alpha&                          c_\alpha\\
     Lc_\alpha + \frac{c_\tau s_\alpha}{c_f}&    - Lc_\alpha - Ls_\alpha& -         Lc_\alpha + Ls_\alpha&- Lc_\alpha - \frac{c_\tau s_\alpha}{c_f}\\ 
     - L  c_\alpha - L  s_\alpha &  \frac{ c_\tau  s_\alpha}{ c_f} - L  c_\alpha&   L  c_\alpha - \frac{ c_\tau  s_\alpha}{ c_f} &   L  c_\alpha + L  s_\alpha \\ 
     \frac{ c_\tau   c_\alpha}{c_f} - L   s_\alpha &- L   s_\alpha -  \frac{ c_\tau   c_\alpha}{c_f} & - L   s_\alpha -  \frac{ c_\tau   c_\alpha}{c_f} & \frac{  c_\tau  c_\alpha}{c_f} - L   s_\alpha \end{matrix}$} \small{\frac{\partial\mathbf{w}}{\partial \alpha}}\right]
    %
     \end{align*}

	\subsection{Power Consumption Analysis}

	%

	To analyze the power consumption of the OmniMorph in comparison with a non-morphing fully actuated platform, let us start by computing the input needed to  enforce the static equilibrium when the platform is hovering with orientation $\mathbf{R}=\mathbf{I}_{3}^{}$\textcolor{black}{; we indicate that input as $\boldsymbol{u}^h_w(\alpha,m)$ and retrieve it by inverting}~\eqref{eq:wrench_map} as 
 \begin{align}
 \boldsymbol{u}^h_w(\alpha,m)=\left(\mathbf{A}(\alpha)\right)^\dagger\begin{bmatrix}
	mg\vE{3}\\\vect{0}
	\end{bmatrix}.    
 \end{align}

 The power consumed by each brushless motor is equal to the scalar product between the motor torque and its angular velocity. Given that in hovering we can approximate the motor torque with the drag moment and that the drag moment is always opposing the angular velocity, we can write the power consumed by the $i$-th brushless motor as $P_{\tau i}(u_{w_{i}})= \|\boldsymbol{\tau}_{d_{i}}\||w_i|=c_\tau|w_i|^3=c_\tau|u_{w_{i}}|^{3/2}$. The total power consumed by all the motors is then 
 \begin{align}
 P_{\tau}(\boldsymbol{u}_w)=\sum_{i=1}^8 P(u_{w_{i}})= c_\tau \sum_{i=1}^8 |u_{w_{i}}|^{3/2}. 
 \end{align}

 Therefore, $P_{\tau}(\boldsymbol{u}^h_w(\alpha,m))$ represents the power consumed at hovering by a platform with a certain mass $m$ and a given tilting angle $\alpha$. 

		\begin{figure}[t]
	    \centering
	    \includegraphics[trim={3cm 8cm 4cm 8cm},clip,width=0.98\columnwidth]{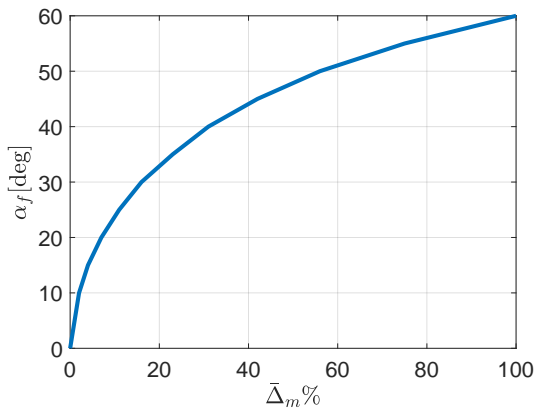}
	    \caption{For a given fixed propeller tilting angle $\alpha_f$, a morphing design is convenient, provided that the relative mass of the tilting mechanism is lower than $\bar{\Delta}_m$ }
	    \label{fig:dm_alpha}
	\end{figure}

 Let us now write the robot's mass as follows: $m=m_0(1+\Delta_m)$, where $\Delta_m$ is the ratio between the mass of the propeller tilting mechanism and $m_0$, which is the mass of all other components; \textcolor{black}{indeed,  $\Delta_m$ is given by $\frac{m-m_0}{m_0}.$} 
 
\begin{figure}[t]
    \centering
\includegraphics[width=\columnwidth, trim={3cm 10cm 3cm 10cm},clip]{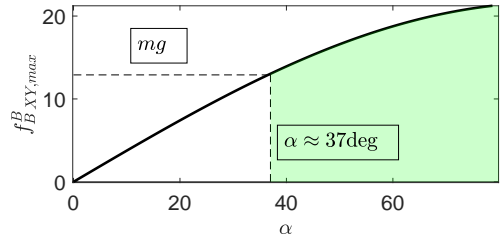}
    \caption{Radius of the maximum inscribed sphere is the feasible force set in Figure \ref{fig:feasible_forces} for different values of $\alpha$.When this is equal to its weight, OmniMorph can sustain its weight in all orientations.}
    \label{fig:fxy}
\end{figure}

\textcolor{black}{Now, consider instead, a \textit{non-morphing} multirotor}, in which a certain level of dexterity is obtained by tilting the propellers by a \textit{fixed} angle $\alpha_f$.  We ask ourselves which is the critical value \textcolor{black}{of $\Delta_m$, call it  $\bar{\Delta}_m,$ }below which a morphing design would be convenient for such a platform. \textcolor{black}{Indeed, it has to be taken into account that the morphing design can efficiently hover with $\alpha=0$ (all propellers parallel to each other, so no thrust is waist into internal forces) but it will have to sustain a larger mass due to the tilting mechanism ($\Delta_m>0$). On the contrary, the non-morphing platform is lighter ($\Delta_m=0$) but due to the propellers'  fixed tilt part of the thrust generated to hover is wasted in internal forces. To understand which is the maximum value of $\Delta_m$ for which the morphing design is the most efficient one, namely to compute $\bar{\Delta}_m$, we numerically solve the following implicit relation between $\Delta_m$ and $\alpha$:} 
$$P_{\tau}(\bm{u}_w^h(0,m_0(1+\bar{\Delta}_m))= P_{\tau}(\bm{u}_w^h(\alpha_f,m_0)).$$
\textcolor{black}{Clearly, $\bar{\Delta}_m$ depends on the tilting angle of the non-morphing design.} 
The values of $\bar{\Delta}_m$ as a function of $\alpha_f$ for the OmniMorph are reported in Figure \ref{fig:dm_alpha}. Intuitively, for $\alpha_f=0,$ namely when no dexterity is required from the fixed-propeller platform, which is then a standard uni-directional platform, one has that $\bar{\Delta}_m=0$. On the other side, for instance, for $\alpha_f$=20\,deg,  the use of a morphing platform is more power-efficient for a weight of the tilting mechanism up to 7\% that of all the other components. For the OmniMorph, $\alpha_f\geq30\rm{deg}$,
which tells us that the morphing platform is more power-efficient than a non-morphing one even for a weight of the tilting mechanism that constitutes up to 16\% of all the other components' weight. Note that the minimum $\alpha_f$ to attain omnidirectionality  depends on the maximum motor speed, the propeller type, and the mass of the vehicle. We are considering here the same bi-directional propellers as in \cite{brescianini_omni-directional_2018} with the same maximum motor speed and a mass $m_0=1.128\rm{kg}$ equal to the mass of the OmniMorph without any tilting mechanism. Considering also the additional mass of the tilting mechanism, the minimum value of $\alpha$ for which OmniMorph is omnidirectional is $\alpha\approx37\rm{deg}$, as shown in Figure \ref{fig:fxy}. Clearly, there is also a maximum value for which OmniMorph loses omnidirectionality. OmniMorph cannot sustain its weight in hovering as soon as $\alpha\geq72\rm{deg}.$ 

\textcolor{black}{The power consumption analysis here presented is at a steady state. The power consumed in a dynamic task by the OmniMorph and by an analogous platform but with fixedly tilted propellers is discussed in Sec. \ref{sec:sim}.}

	\section{Control}\label{Sec:control}
%

This section introduces a control law that allows Omni\-Morph to follow a desired 6D trajectory switching between the underactuated and the fully actuated configurations to account for the minimization of trajectory tracking error and the minimization of the input $\bm{u}_w$.

Let us define $\ddot{\boldsymbol{q}}=[
\ddot{\vect{p}}^\top,\ 
\dot{\boldsymbol{\omega}}^\top]^\top$. 
Given a reference trajectory for the UAV, indicated through the superscript $(\cdot)^d$, we  
compute the reference acceleration of the robot, call it $\ddot{\boldsymbol{q}}^r$, using a PD feedback control plus a feedforward term.  The  following desired input acceleration is obtained:
\begin{equation}
\ddot{\boldsymbol{q}}^r=\begin{bmatrix}\ddpD +  \gainK{p1}\dot{\bm{e}}_p+ \gainK{p2} \bm{e}_p\\
\angAccD +\gainK{\omega1} \bm{e}_\omega +\gainK{\omega2}\errR,
\end{bmatrix}
    \label{eq:F-L}
\end{equation}
where ${\errR=\frac{1}{2}[\rot^\top\rotMatD-{\rotMatD}^\top\rot]_\vee}$, with $_\vee$ from $so(3)$ to $\mathbb{R}^3$ being the inverse of the hat map \cite{lee_geometric_2010}, ${\bm{e}}_p=\pD-\pos$, and ${\bm{e}_\omega=\angVelD-\angVel}$.

We now design an inner control loop to track ${\ddot{\boldsymbol{q}}^r}$ thanks to suitable inputs ${\alpha^\ast}$ and ${\boldsymbol{u}_w^\ast},$ chosen as the solution to:

\begin{mini}|s|
{\alpha, \boldsymbol{u}_{w}, \ddot{\boldsymbol{q}}}{J_1 + J_2 + J_3}
{}{}
\addConstraint{\inertia\ddot{\boldsymbol{q}}=\gravityCoriolis+{\mathbf{J}}_R\mathbf{F}(\alpha)\boldsymbol{u}_{w}}
\addConstraint{\mathbf{C}\boldsymbol{u}_{w}<\bm{b}}
\addConstraint{-\epsilon_\alpha\leq\alpha-\alpha^\ast_{k-1}\leq \epsilon_\alpha}
\label{eq:min}
\end{mini}
where the cost function is composed of three terms: ${J_1=||\boldsymbol{u}_{w}||^2_{\matr{W}_1}}$ is to minimize the norm of the input, ${J_2= ||\ddot{\boldsymbol{q}}^r-\ddot{\boldsymbol{q}}||_{\matr{W}_2}^2}$ to ensure tracking of the desired trajectory, and ${J_3=||\boldsymbol{u}_{w}-{\boldsymbol{u}^*}_{w,k-1}||^2_{\matr{W}_3}}$ to minimize the propeller spinning accelerations.  The quantities ${\alpha^*}_{k-1}$ and ${\boldsymbol{u}^*}_{w,k-1}$ are the solution of the optimization problem \eqref{eq:min} at the previous time step. $||\cdot||_{W_i}$ is the 2-norm weighted by the positive definite weight matrix $\matr{W}_i$. Alternatively, if the bounds on the propeller accelerations are identified, as in \cite{bicego2020nonlinear},  $J_3$ could be also expressed as a unilateral constraint.

The equality constraint in \eqref{eq:min} is the system's dynamics, 
where we have indicated  the inertia matrix as 
${\inertia=\begin{bmatrix}
m \mathbf{I}_{3}^{} & \boldsymbol{0}_{3}^{} \\
\boldsymbol{0}_{3}^{} & \mathbf{J}
\end{bmatrix}},$ 
%
 the vector of the gravity and Coriolis terms as  ${\gravityCoriolis=\begin{bmatrix}
-m g \vect{e}_{3}^{} \\
-\boldsymbol{\omega} \times    \mathbf{J} \boldsymbol{\omega}	
\end{bmatrix}},$ and $\mathbf{J}_R=\begin{bmatrix}
\mathbf{R} & \boldsymbol{0}_{3}^{} \\
\boldsymbol{0}_{3}^{} & \mathbf{I}_{3}^{}
\end{bmatrix}$; the second, unilateral, constraint in  \eqref{eq:min} expresses the input constraint, where $\mathbf{C}, \bm{b}$ are properly defined constant quantities; the last constraint is on the rate of change of $\alpha$. The maximum rate of change $\epsilon_\alpha$ is here defined as symmetric but in the general case they may also be non-symmetric and the analysis still holds. 
The term $J_1$ has an equivalent effect to minimizing the norm of the drag moments, as it differs from the input norm by a constant coefficient.  

In order to solve problem \eqref{eq:min}, which is not a QP problem because the first constraint is nonlinear in the optimization variables, we proceed as follows. Consider the following problem for a fixed value of $\alpha$, indicated as $\bar{\alpha}.$
\begin{mini}|s|
{\boldsymbol{u}_{w}, \ddot{\boldsymbol{q}}}{J_1 + J_2 + J_3}
{}{}
\addConstraint{\inertia\ddot{\boldsymbol{q}}=\gravityCoriolis+{\mathbf{J}}_R\mathbf{F}(\bar{{\alpha}})\boldsymbol{u}_{w}}
\addConstraint{\mathbf{C}\boldsymbol{u}_{w}<b}
\label{eq:min2}
\end{mini}
 Problem \eqref{eq:min2} is now a QP problem as $\alpha$ is not an optimization variable anymore, and, as a consequence, the bilateral constraint is linear in the optimization variables.  
 
\begin{figure}[t]
     \centering \includegraphics[width=\columnwidth]{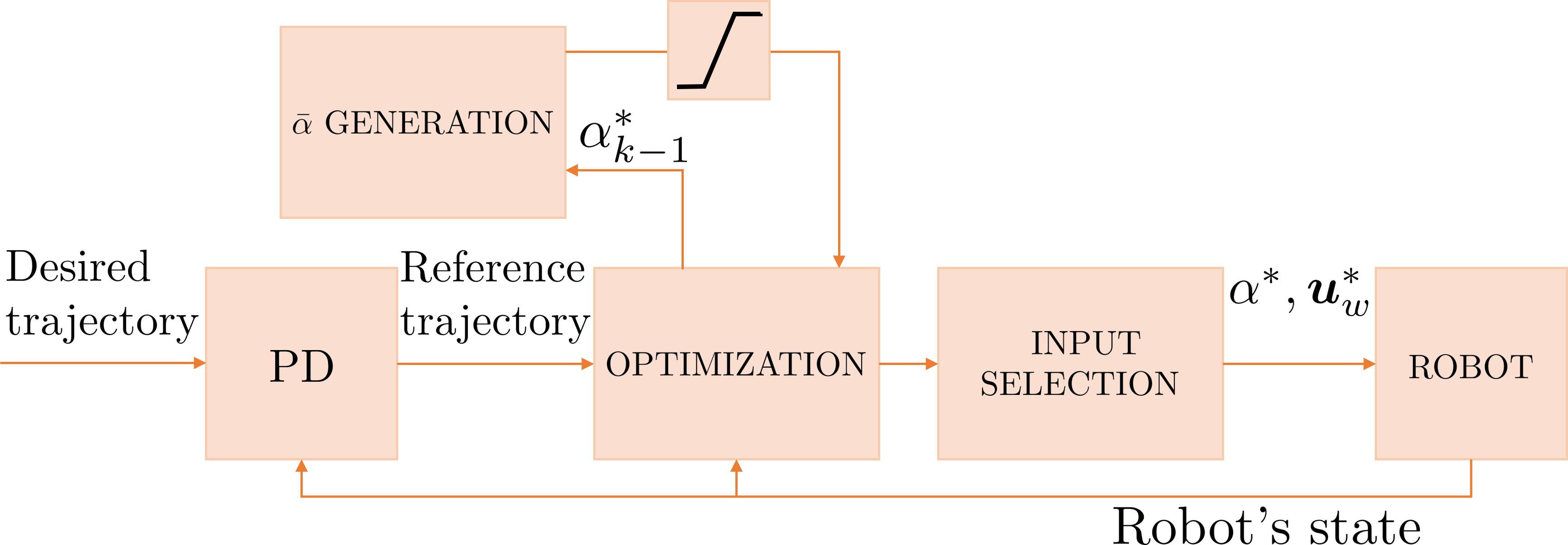}
     \caption{\textcolor{black}{Block diagram of the control method.}}
     \label{fig:ctrl_scheme}
 \end{figure}

 At each time step~\eqref{eq:min2} is solved three times for three different values of $\bar{{\alpha}}=\{{\alpha^\ast}_{k-1}- \epsilon_\alpha,\,{{\alpha}^\ast}_{k-1},\, {{\alpha}^\ast}_{k-1}+ \epsilon_\alpha\}.$ Hence, at each time step, the solutions $\alpha^\ast$ and $\bm{u}_w^\ast$ that correspond to the lowest value of the cost function are selected. A schematic representation of the proposed control scheme is in Figure \ref{fig:ctrl_scheme}.
\textcolor{black}{The block that feeds the optimization with the three different values of $\alpha$ also performs a saturation to make sure that the values of $\alpha$ provided by the controller comply with the platform kinematics constraints. Especially, the propellers are allowed to assume values between 0 and 60 deg. We saw that the  theoretical maximum value of $\alpha$ for OmniMorph to sustain its weight in any configuration was about 70 deg; however, we decided to be more conservative than the theoretical upper bound to keep always enough thrust in the vertical direction to account for model uncertainties, rejection of external disturbances, low battery levels, etc.}
 
 In principle, the selection of the parameter $\bar{\alpha}\in[{\alpha}^\ast_{k-1}-\epsilon_\alpha, {\alpha}^\ast_{k-1}+\epsilon_\alpha]$  may be done following different methods. Here we proposed a simple method that allows keeping the value of the tilting angle constant or changing it in the two possible directions accounting for the bound on the variation of $\alpha$.  Other approaches such as an exhaustive search or more efficient sampling algorithms are possibilities, and their assessment is left for future work. 
 

Note that, by tuning the weights $\matr{W}_i$, the control input negotiates between low input effort and low tracking error. We recall that, thanks to its redundancy, the robot is able to follow a certain desired trajectory with different values of the control inputs. 
	\section{Results}
 \subsection{Simulations}\label{sec:sim}

Numerical simulations have been carried out using a \textcolor{black}{Unified Robot Description Format (URDF)} description of the OmniMorph and \textcolor{black}{ordinary differential equation (ODE)} physics engine in Gazebo. The control software has been implemented in Matlab-Simulink. The interface between Matlab and Gazebo is managed by a Gazebo-genom3 plugin\footnote{\url{https://git.openrobots.org/projects/mrsim-gazebo}}. A Simulink s-function embedding qpOases QP solver\footnote{\url{https://github.com/coin-or/qpOASES}} has been used for the optimization problem.  
\begin{figure}[t]\label{fig:Gazebo_res}
\centering
    \subfloat[][Desired position  \label{fig:trajp}]{\includegraphics[trim={0cm 11cm 0cm 11cm},clip,width=0.9\columnwidth]{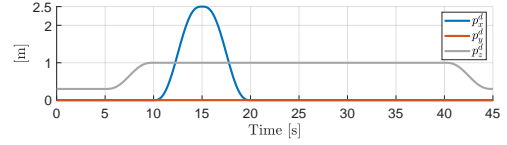}}\\
   \subfloat[][Desired attitude expressed as a rotation around $\mathbf{x}_W$ \label{fig:trajR}]{ 
   \includegraphics[trim={0cm 11cm 0cm 11cm},clip,width=0.9\columnwidth]{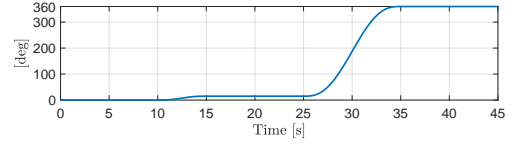}}\\
    \subfloat[][Case A: in orange line; Case B: blue line 
    \label{fig:sim_results}]{\includegraphics[trim={1cm 4.5cm 2cm 3.5cm},clip,width=0.97\columnwidth]{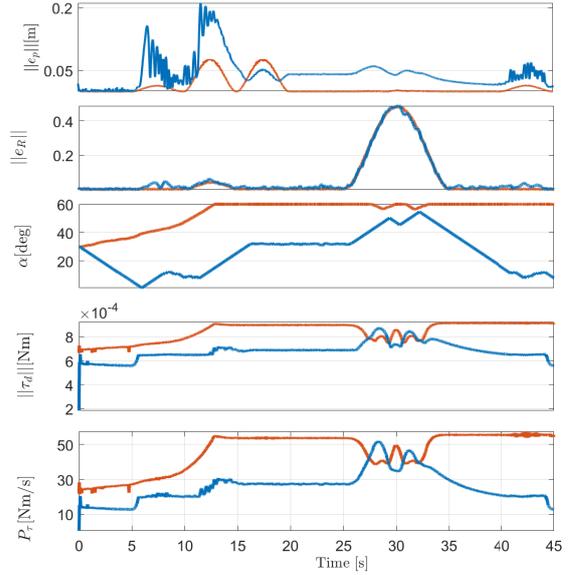}}\caption{\textcolor{black}{Desired trajectory (\ref{fig:trajp} and \ref{fig:trajR}) and results (\ref{fig:sim_results}) of the Gazebo simulations.}}\end{figure}
The mass and inertia of the robot are $m=1.3150\,\rm{kg}$ and ${\mathbf{J}=\rm{diag}( 1.16\cdot  10^{-2},1.13\cdot  10^{-2}, 1.13\cdot  10^{-2})\,\rm{Nms^2}}$
The other parameters are
\begin{itemize}
    \item ${\gainK{p1}=\rm{diag}(30,30,30)}\,\rm{s^{-1}}$,  \item $\gainK{p2}=\rm{diag}(300,300,300)\,\rm{s^{-2}}$,  
    \item${\gainK{\omega1}=\rm{diag}(40,40,40)}\rm{s^{-1}}$, ${\gainK{\omega2}=\rm{diag}(100,100,100)}\,\rm{s^{-2}}$, 
    \item ${\matr{W}_3=10^{-5}\mathbf{I}_8}$. 
\end{itemize}
The tilting angle $\alpha$ is saturated at 60 deg, a value sufficient to reach the omnidirectionality, as it will be in the real robot. 
\textcolor{black}{A PD controller has been used to generate the reference trajectory of the robot's CoM, but any other suitable controller could be used for that purpose, e.g., based on dynamic inversion, etc. Indeed, the main contribution of this work from a control point of view is the optimization, and PD gains are not of primary importance for showing the work's contribution. Fine gain tuning, although important for performance enhancement, is, thus, considered out of the scope of this work. The PD gains have been handcrafted and set once and for all. All the following results are purposely obtained with the same PD controller gains, to highlight the contribution of the optimization.  Advanced methods for automatic gain tuning based, e.g., on neural networks\cite{lin2001line}, reinforcement learning \cite{el2013application}, or stochastic optimization \cite{kose2023simultaneous} have been proposed in the literature, and they can be considered for refining the performance of the controlled system in future work.}

We show two different simulation scenarios, let us refer to them as Case A and Case B, in which the robot's desired trajectory is the same, but the optimization gains are different. Especially, the trajectory includes moving upwards/downwards, translating with a tilted attitude, and rotating by 360 degrees on the spot\textemdash see Figures~\ref{fig:trajp} and \ref{fig:trajR}. It has been chosen to cover the whole spectrum of the actuation modes: the vertical motion with constant upward orientation is feasible in the underactuated mode, the lateral translations with constant upward orientation require full actuation, and the full 360\,deg rotation on the spot about a horizontal axis requires omnidirectionality.

What distinguishes the two scenarios is that Case A is characterized by a lower weight on the input, $\matr{W}_1$, and a higher weight on the tracking error, $\matr{W}_2$, than  Case B.

The optimization weights for Case A are $\matr{W}_1=10^{-8}\mathbf{I}_8$ and ${\matr{W}_2=\rm{diag}(3\cdot  10^{6}, 3\cdot  10^{6}, 3\cdot  10^{6}, 10^3, 10^3, 10^3)}$. The optimization weights for Case B are $\matr{W}_1=10^{-5}\mathbf{I}_8$ and $\matr{W}_2=\rm{diag}(3\cdot 10^{4}, 3 \cdot10^{4}, 3\cdot10^{4}, 10, 10, 10)$.

In Figure~\ref{fig:sim_results}, we report a comparison between the results of the two simulation scenarios. One can appreciate as in Case A the robot performs better in terms of tracking (the average position and attitude errors are, respectively, $0.011\,\rm{m}$ and $0.062$), at the expenses of a higher motor torque; the tilting angle is increased and kept at the highest value along the task execution. On the contrary, in Case B, the errors are higher (mean square position and attitude errors are, respectively,  $0.044\,\rm{m}$ and $0.068$), but the motor torque is, as expected, lower, and the propeller tilting angle is decreased as soon as the robot only moves vertically (as soon as the underactuated mode is enough to perform the desired motion). The efficiency is higher in Case B with the value of energy related to the drag moment equal to $1162.3\,\rm{J}$ against $2080.7\,\rm{J}$ in Case A. 
\begin{figure}[t]
\centering
 \includegraphics[trim={2cm 10.5cm 3cm 6cm},clip,width=0.9\columnwidth]{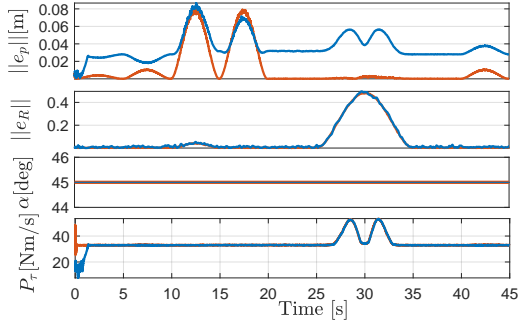}
 \caption{\textcolor{black}{Results of the Gazebo simulations in Case A (red) and Case B (blue) for the propeller tilting angle fixed at $\alpha=45\,\rm{deg}$. The power consumption remains almost identical in the two cases.}}\label{fig:Gazebo_res_45deg}
 \end{figure}
\textcolor{black}{Note that, in a trajectory tracking task like the one shown here, there is another source of power consumption besides the power consumed by the drag: it is the power needed to accelerate the propellers. If $\text{d}E$ is the instantaneous variation of the kinetic energy of all the eight propellers, the instantaneous power is computed as $\frac{\text{d} E}{\text{d}t}$. Considering the mass of each propeller equal to $0.01$kg,  the total energy consumed to accelerate all the propellers is equal to $65.5\, \rm{J}$ in Case A and $225.3\,\rm{J}$ in Case B.}

\textcolor{black}{This suggests that only a minor part of the energy is spent to accelerate the propellers, and Case B remains the most efficient scenario. One can act on $\matr{W}_3$ to reduce the propeller acceleration. We carried out the same simulation as in Case B but increasing the weight on the propeller acceleration to $\matr{W}_3=10^{-4}$; we obtained a total consumed energy to accelerate the propellers equal to $87.7\, \rm{J}$. The total corresponding tracking errors were comparable to Case B: $0.0412\,\rm{cm}$ in position and $0.063$ in attitude.}

\textcolor{black}{
\subsubsection{Simulations at a fixed $\alpha$} We performed simulations in Case A and Case B with a fixed value of $\alpha=45$ deg\textemdash see Fig. \ref{fig:Gazebo_res_45deg}. The total consumed energy is around $1400\,\rm{J}$ in both cases, namely in between the energy consumed by OmniMorph to perform an analogous task in Case A and Case B. The ratio between the energy consumed by the fixed-propeller robot and OmniMorph ultimately depends on the specific task, e.g., on the portion of vertical motion and hovering, not requiring large values of $\alpha$.}

\textcolor{black}{However, it is worth noting that fixedly tilted propellers seem to offer less room to tune the efficiency, as the power consumption corresponding to the same reference trajectory remains basically the same for Cases A and B. Instead, OmniMorph offers the possibility to exploit the additional degree of freedom to reach the desired balance between efficiency 
 and performance.}

 \begin{figure}[t]
     \centering
\includegraphics[width=\columnwidth]{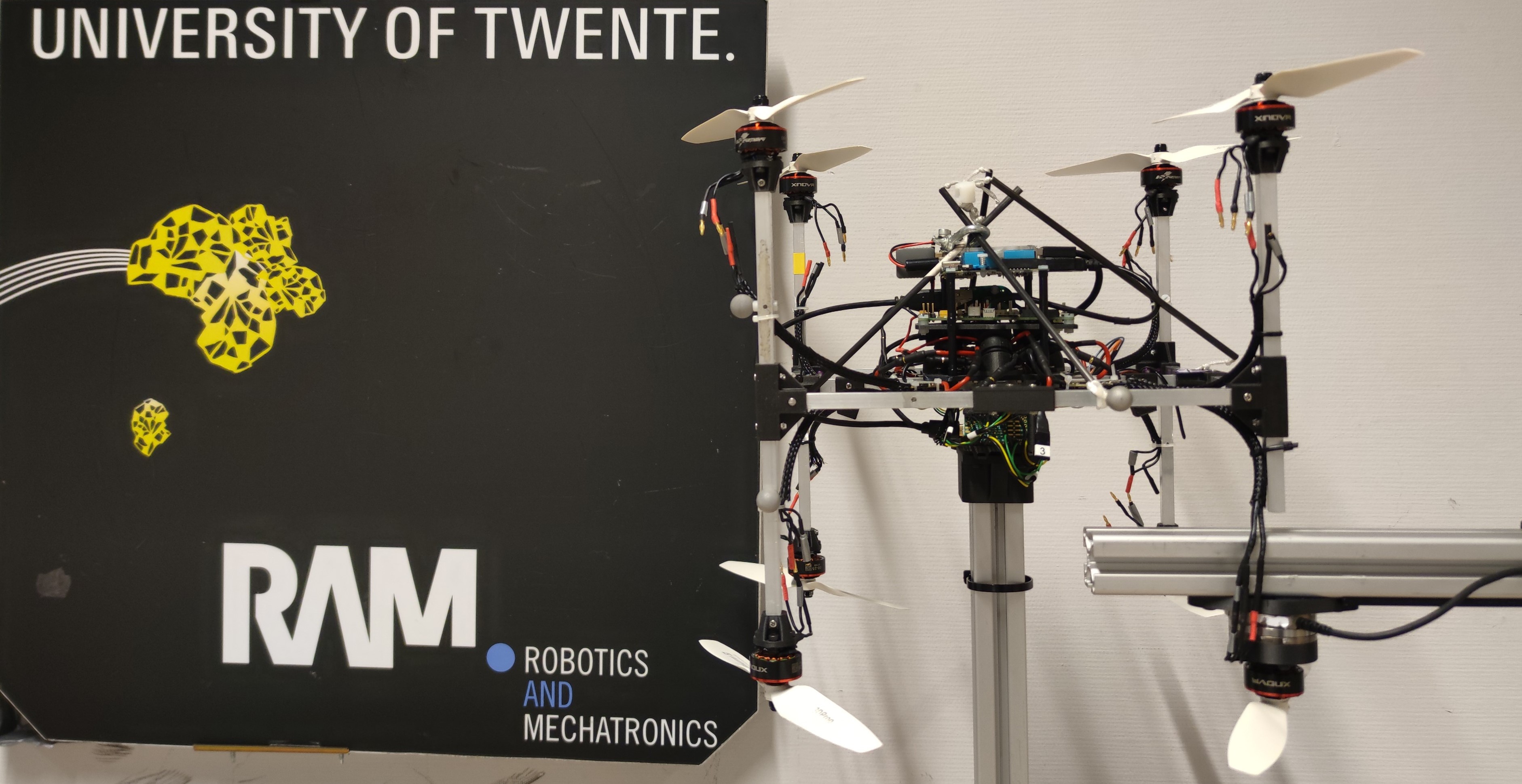}
     \caption{\textcolor{black}{Experimental setup for ${\alpha=0}$. }}
     \label{fig:exp_setup}
 \end{figure}
 
\begin{figure}[t]
     \centering
     \includegraphics[trim={1cm 8.5cm 3cm 8cm},clip,width=0.35\textwidth]{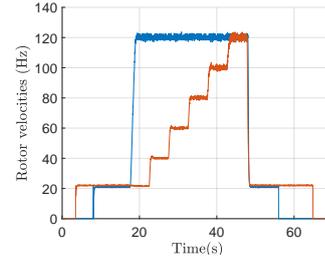}
     \caption{\textcolor{black}{Profiles of the propellers' velocities during the tests. In blue, the constant velocity of the propeller attached to the force sensor; in red the velocity of the others.}}
     \label{fig:exp_vel}
 \end{figure}

\begin{figure}[t]
    \centering
\includegraphics[trim={3cm 9cm 3cm 8.5cm},clip,width=0.9\columnwidth]{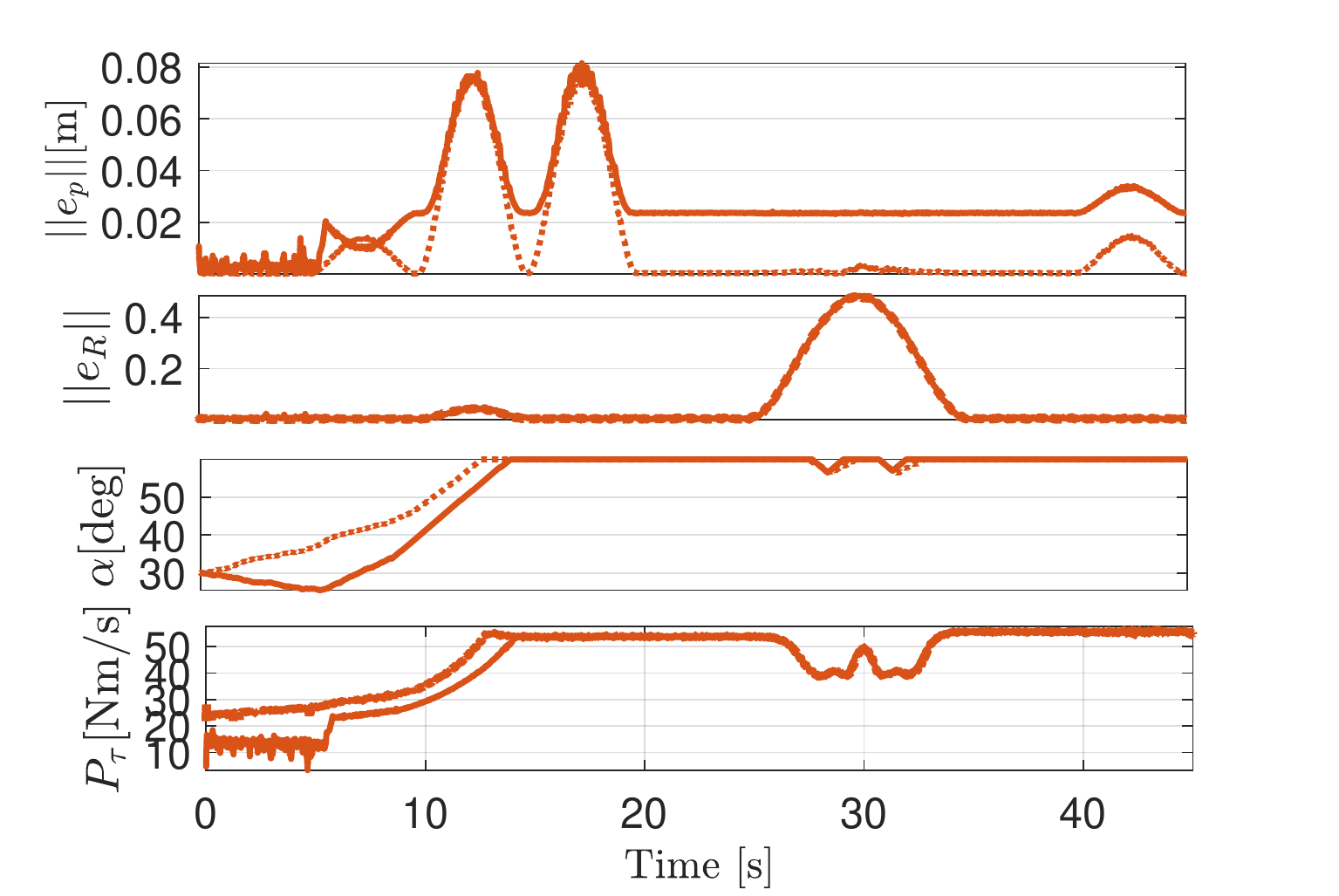}\\
\includegraphics[trim={3cm 9cm 3cm 8cm},clip,width=0.9\columnwidth]{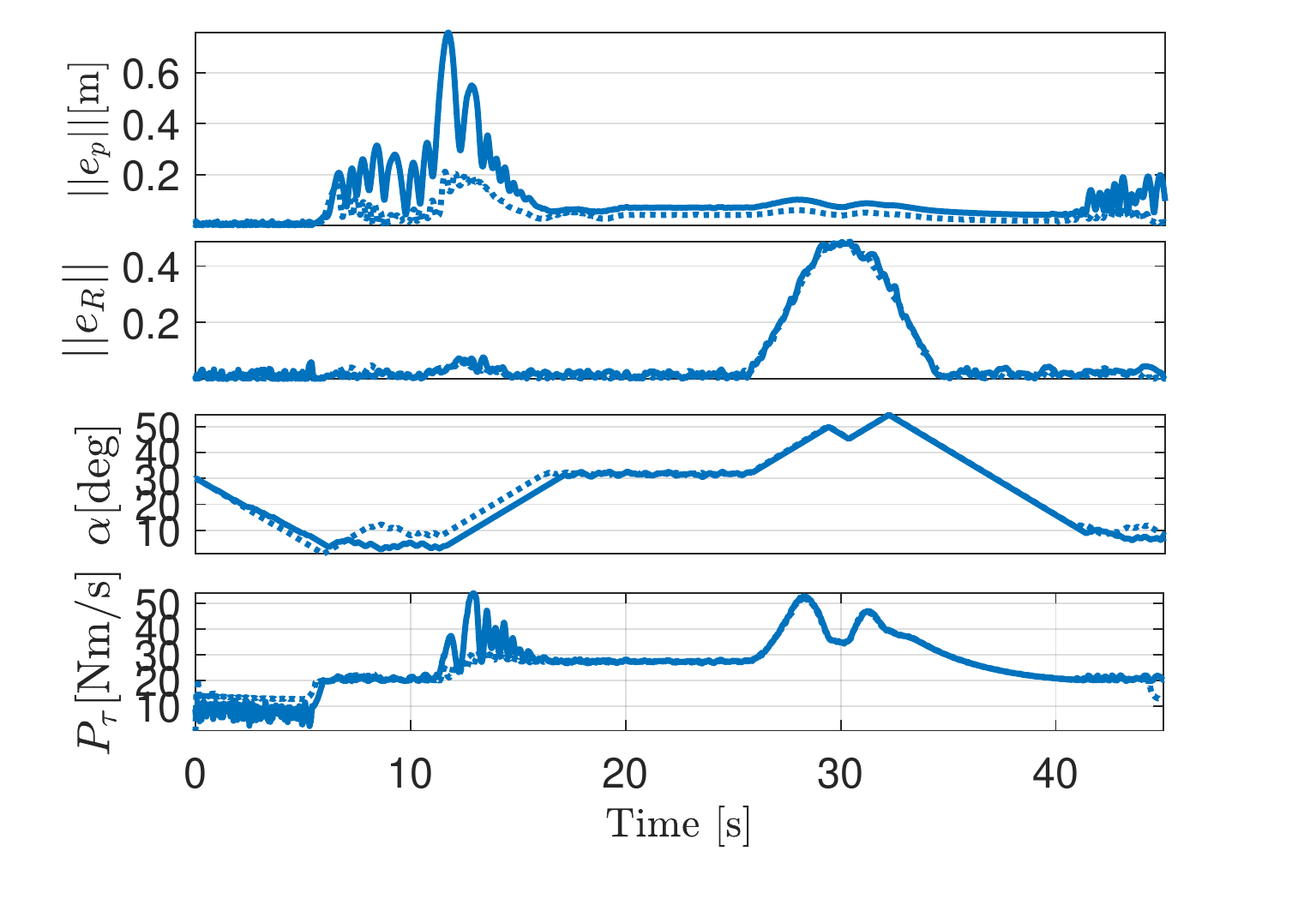}
    \caption{\textcolor{black}{Top: results for Case A with actual $c_f$ lower than the nominal one by $35\%$. Bottom: results for Case B with actual $c_f$ lower than the nominal one by $30\%$. The results of the ideal cases are superimposed in dotted lines.}}
  \label{fig:cf_decreasedAB}
\end{figure}

 \textcolor{black}{\subsection{Propellers' mutual aerodynamic interference.}\subsubsection{Experiments}
 In this section, we present preliminary results to experimentally assess the propellers' mutual aerodynamic interference in the OmniMorph prototype. A picture of the experimental setup is in Figure \ref{fig:exp_setup}. A prototype of OmniMorph without the tilting mechanism is attached to a fixed structure. A propeller was fixed to an ATI Mini40\footnote{\url{https://www.ati-ia.com/products/ft/ft_models.aspx?id=Mini40}} force-torque sensor at the same location it would have on the OmniMorph body. The propeller measured by the force torque sensor was commanded to spin at a constant rate of 120Hz. The remaining 7 propellers were accelerated in a stepwise fashion with 20Hz increments to a maximum propeller speed of 120Hz, as reported in Figure \ref{fig:exp_vel}. Electronic speed controllers Kiss Racing 32A\footnote{\url{https://www.getfpv.com/kiss-32a-32bit-esc-2-6s.html}} were used. 
 The effect of aerodynamic interference was estimated by measuring the change in force on the single propeller due to the increasing speed of the remaining propellers.
 A propeller in the lower part has been purposely chosen to consider the worst-case scenario as it gets invested by the downward airflow of the top propeller. Two values of $\alpha=[0, 50]$ deg have been tested by fixing the propellers with custom 3D-printed parts.} 
 \begin{figure}[t]
     \centering
      \includegraphics[trim={3.5cm 8.5cm 3.5cm 9.5cm},clip,width=0.47\columnwidth]{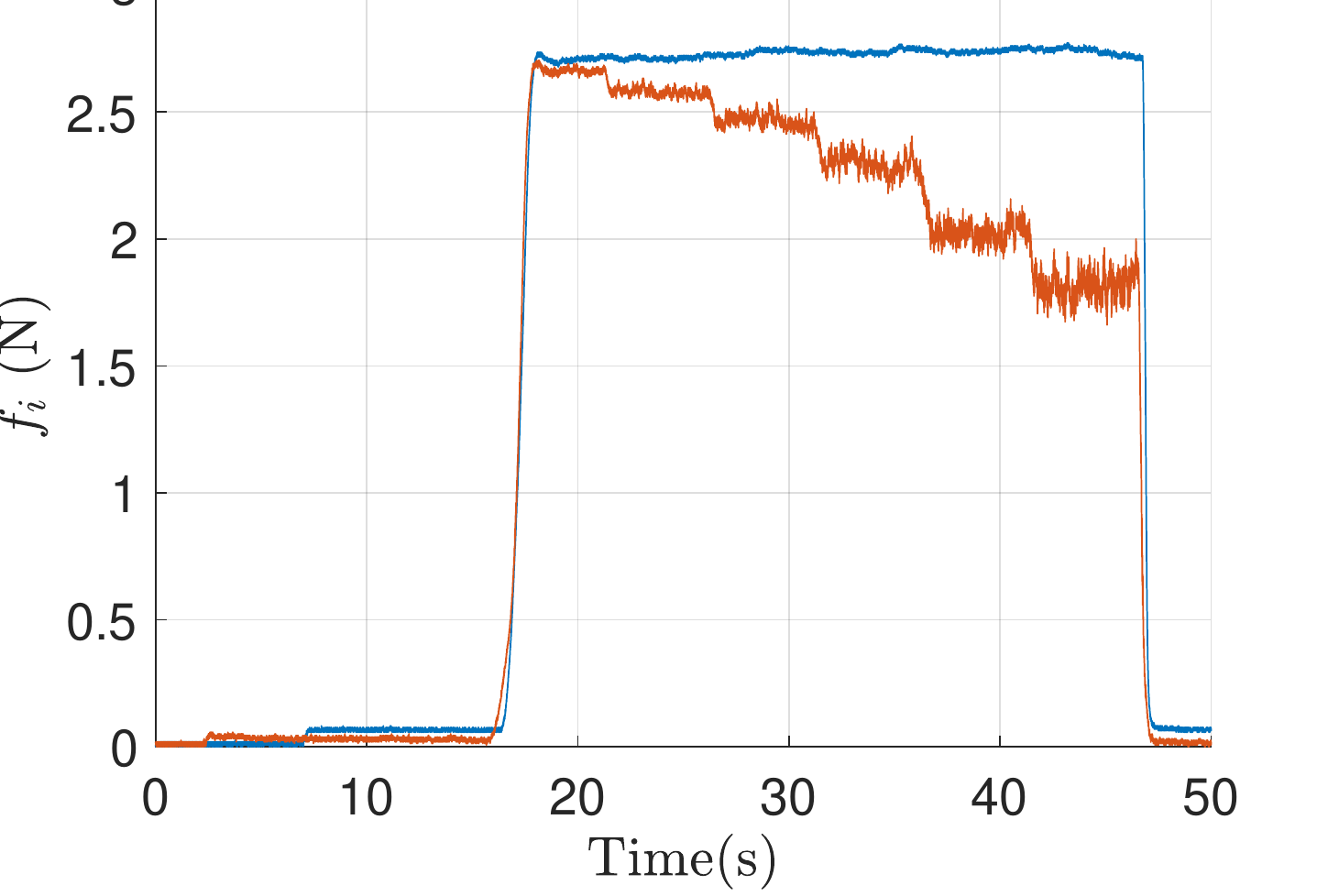} \includegraphics[trim={3.5cm 8.5cm 3.5cm 9.5cm},clip,width=0.47\columnwidth]{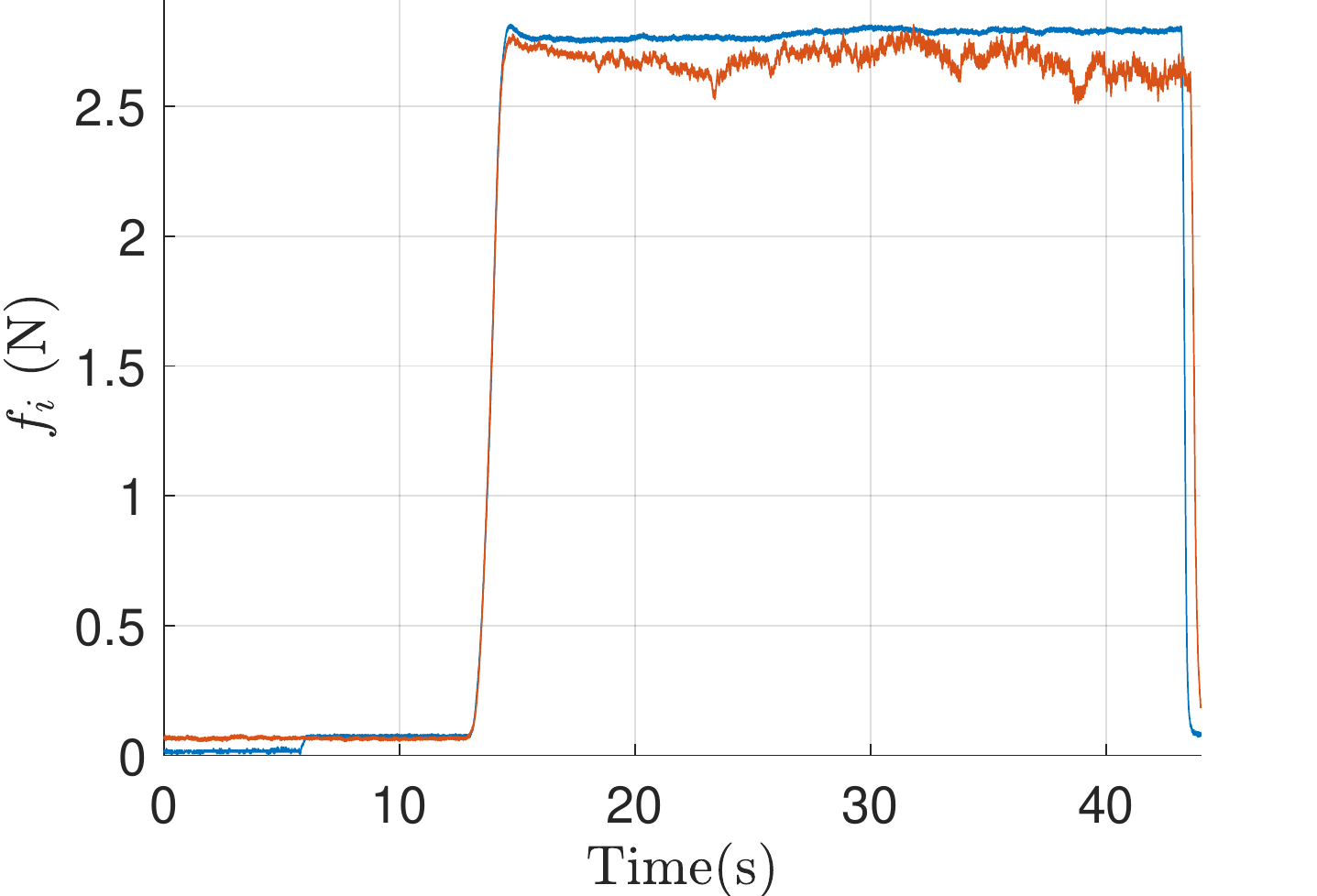}
     \caption{\textcolor{black}{The blue values are the force sensor measures when the propeller attached to the sensor is the only one spinning. The red values are the measures when the other propellers spin as in Figure \ref{fig:exp_vel}}. $\alpha=0$ on the left; $\alpha=50$ deg on the right. }
     \label{fig:exp_force}
 \end{figure}
 \textcolor{black}{It emerges that, as intuitively expected, the aerodynamic interference is higher when $\alpha=0$, as the entire airflow of the top propellers is directed downwards toward the bottom ones. In that case, the measured force generated by the individual propeller goes from $2.7$ N  to about $1.8$ N, i.e., it decreases by a maximum of about $33\%$  of its nominal value. On the other hand, the maximum decrease when $\alpha=50\%$ is around $7\%$. These results are reported in Figure \ref{fig:exp_force}. 
 The non-trivial computation of the perturbed allocation matrix as a function of all propellers' individual speed and of the angle $\alpha$ is left as future work. However, in the next section, we show simulation results to assess the robustness of the proposed controller to aerodynamic interferences.} 
  \textcolor{black}{\subsubsection{Simulations with  diminished propeller performance }
Based on the previous results, in this section, we show two simulations to assess the robustness of the proposed controller against decreased propeller performance. Especially, the actual $c_f$ of OmniMorph is lower than the nominal one known by the controller.
In reality, the amount varies for each propeller depending on the others' spinning velocity and $\alpha$. The assessment of an update low for $c_f$ based on experimental data is left for future work. As a worst-case scenario, we considered in the simulations that the actual thrust coefficient is constant and $35\%$ lower than the nominal one. While the platform successfully carries out the task in Case A, although with a slightly higher tracking error than in the ideal case, in Case B the robot fails to recover from high tracking errors seemingly due to the scarce dexterity in the horizontal directions caused by low values of $\alpha$; however, Case B is still successful when $c_f$ is $30\%$ lower than the nominal one. That suggests that, in the presence of such a mismatch between the model and the system, high priority should be given to the tracking performance to improve robustness. However, we simulated the worst-case scenario in which the mismatch is constantly high and is the\textit{ same for all propellers}, while in the real platform, the top propellers would likely suffer from considerably lower interference.
Figure~\ref{fig:cf_decreasedAB} collects the results in the two aforementioned cases. The position error is the most affected in both cases. Recordings of the simulations are provided in the attached multimedia material.}


	\section{Conclusions}\label{sec:conclusion}
	This work presented the design concept of  OmniMorph, a novel morphing multirotor that can range between omnidirectionality and underactuation thanks to actively tilting propellers. The design leverages one single servomotor to synchronously tilt all the propellers, thus reducing the mechanical complexity and the additional payload. The dynamics model was presented, and the actuation properties depending on the propeller tilting angles were studied. Hence, a controller was proposed to negotiate between input effort and tracking performance. Simulations in a realistic Gazebo environment were presented. \textcolor{black}{A preliminary prototype was built and tests were carried out to assess the propeller aerodynamic interferences on the real prototype, and additional simulation results to assess the robustness against them were shown.}
	
	In the future, experiments on the real prototype will be carried out, and predictive controllers will be tested. \textcolor{black}{Computing the online changes of the allocation matrix due to aerodynamic interferences depending on all individual propeller speeds and on $\alpha$, and incorporating such information on the optimal control is left as interesting future work. Increasing the dimensions of the prototype to allow better spacing of the propellers will also be considered.}  Equipping the OmniMorph with physical interaction capabilities in an interesting future direction. 
 \section{Declarations}
%
\begin{itemize}
\item{\textbf{Acknowledgments}}
The authors thank Gianluca Corsini for contributing to the simulator and Aaron Saini for the code to plot the feasible force sets.
\item{\textbf{Funding}} This work was partially funded by European Commission projects Horizon2020 AUTOASSESS (101120732) and MSCA project Flyflic (101059875).
\item{\textbf{Conflicts of interest}}
The authors have no relevant financial or non-financial interests to disclose.
\item{\textbf{Code or data availability}} 
Data sets generated during the current study are available from the corresponding author on reasonable request.
\item{\textbf{Authors' Contributions}} 
Y. Aboudorra contributed to deriving the modeling, the control law, designing the simulator, and writing the manuscript, as well as to the background literature research.
C. Gabellieri contributed to the theoretical analysis, controller implementation, and manuscript writing, as well as to the background literature research; she worked on the simulation results and the multimedia files.
R. Brantjes contributed to testing the prototype and to the manuscript writing.
Q. Sablé contributed to the OmniMorph design and to building and testing the prototype.
A. Franchi is the initiator of the OmniMorph research idea, he did the fundraising and is the PI of the research project. He contributed to the modeling and the control law, and to the manuscript writing, as well as to the background literature research.
\item{\textbf{Ethics approval}} 
Not Applicable
\item{\textbf{Consent to participate}} 
Not Applicable
\item{\textbf{Consent for publication}} 
Not Applicable
\end{itemize}	
	
\bibliography{references.bib, bibCustom.bib}

\begin{thebibliography}{10}

\bibitem{ollero_past2022}
Anibal Ollero, Marco Tognon, Alejandro Suarez, Dongjun Lee, and Antonio
  Franchi.
\newblock Past, {Present}, and {Future} of {Aerial} {Robotic} {Manipulators}.
\newblock {\em IEEE Transactions on Robotics}, 38(1):626--645, February 2022.

\bibitem{tognon_physical2021}
Marco Tognon, Rachid Alami, and Bruno Siciliano.
\newblock Physical {Human}-{Robot} {Interaction} {With} a {Tethered} {Aerial}
  {Vehicle}: {Application} to a {Force}-{Based} {Human} {Guiding} {Problem}.
\newblock {\em IEEE Transactions on Robotics}, 37(3):723--734, June 2021.

\bibitem{tognon2019truly}
Marco Tognon, Hermes A~Tello Ch{\'a}vez, Enrico Gasparin, Quentin Sabl{\'e},
  Davide Bicego, Anthony Mallet, Marc Lany, Gilles Santi, Bernard Revaz, Juan
  Cort{\'e}s, et~al.
\newblock A truly-redundant aerial manipulator system with application to
  push-and-slide inspection in industrial plants.
\newblock {\em IEEE Robotics and Automation Letters}, 4(2):1846--1851, 2019.

\bibitem{gabellieri2023equilibria}
Chiara Gabellieri, Marco Tognon, Dario Sanalitro, and Antonio Franchi.
\newblock Equilibria, stability, and sensitivity for the aerial suspended beam
  robotic system subject to model uncertainty.
\newblock {\em arXiv preprint arXiv:2302.07031}, 2023.

\bibitem{corsini2022nonlinear}
Gianluca Corsini, Martin Jacquet, Hemjyoti Das, Amr Afifi, Daniel Sidobre, and
  Antonio Franchi.
\newblock Nonlinear model predictive control for human-robot handover with
  application to the aerial case.
\newblock In {\em 2022 IEEE/RSJ International Conference on Intelligent Robots
  and Systems (IROS)}, pages 7597--7604. IEEE, 2022.

\bibitem{hamandi_design_2021}
Mahmoud Hamandi, Federico Usai, Quentin Sablé, Nicolas Staub, Marco Tognon,
  and Antonio Franchi.
\newblock Design of multirotor aerial vehicles: {A} taxonomy based on input
  allocation.
\newblock {\em The International Journal of Robotics Research},
  40(8-9):1015--1044, August 2021.
\newblock Publisher: SAGE Publications Ltd STM.

\bibitem{kose2023simultaneous}
Oguz Kose and Tugrul Oktay.
\newblock Simultaneous design of morphing hexarotor and autopilot system by
  using deep neural network and spsa.
\newblock {\em Aircraft Engineering and Aerospace Technology}, 95(6):939--949,
  2023.

\bibitem{csahin2022simultaneous}
H{\"u}seyin {\c{S}}ahin, Oguz Kose, and Tugrul Oktay.
\newblock Simultaneous autonomous system and powerplant design for morphing
  quadrotors.
\newblock {\em Aircraft Engineering and Aerospace Technology},
  94(8):1228--1241, 2022.

\bibitem{kose2020simultaneous}
Oguz Kose and Tugrul Oktay.
\newblock Simultaneous quadrotor autopilot system and collective morphing
  system design.
\newblock {\em Aircraft Engineering and Aerospace Technology},
  92(7):1093--1100, 2020.

\bibitem{kose2023simultaneous2}
Oguz Kose, Tugrul Oktay, and Enes {\"O}zen.
\newblock Simultaneous arm morphing quadcopter and autonomous flight system
  design.
\newblock {\em Aircraft Engineering and Aerospace Technology}, 2023.

\bibitem{hamandi_understanding_2021}
Mahmoud Hamandi, Quentin Sable, Marco Tognon, and Antonio Franchi.
\newblock Understanding the omnidirectional capability of a generic multi-rotor
  aerial vehicle.
\newblock In {\em 2021 {Aerial} {Robotic} {Systems} {Physically} {Interacting}
  with the {Environment} ({AIRPHARO})}, pages 1--6, October 2021.

\bibitem{hamandi_omni-plus-seven_2020}
Mahmoud Hamandi, Kapil Sawant, Marco Tognon, and Antonio Franchi.
\newblock Omni-{Plus}-{Seven} ({O7}+): {An} {Omnidirectional} {Aerial}
  {Prototype} with a {Minimal} {Number} of {Unidirectional} {Thrusters}.
\newblock In {\em 2020 {International} {Conference} on {Unmanned} {Aircraft}
  {Systems} ({ICUAS})}, pages 754--761, September 2020.
\newblock ISSN: 2575-7296.

\bibitem{brescianini_design_2016}
Dario Brescianini and Raffaello D'Andrea.
\newblock Design, modeling and control of an omni-directional aerial vehicle.
\newblock In {\em 2016 {IEEE} {International} {Conference} on {Robotics} and
  {Automation} ({ICRA})}, pages 3261--3266, May 2016.

\bibitem{park_design2016}
Sangyul Park, Jongbeom Her, Juhyeok Kim, and Dongjun Lee.
\newblock Design, modeling and control of omni-directional aerial robot.
\newblock In {\em 2016 {IEEE}/{RSJ} {International} {Conference} on
  {Intelligent} {Robots} and {Systems} ({IROS})}, pages 1570--1575, 2016.
\newblock ISSN: 2153-0866.

\bibitem{park_odar2018}
Sangyul Park, Jeongseob Lee, Joonmo Ahn, Myungsin Kim, Jongbeom Her, Gi-Hun
  Yang, and Dongjun Lee.
\newblock {ODAR}: {Aerial} {Manipulation} {Platform} {Enabling}
  {Omnidirectional} {Wrench} {Generation}.
\newblock {\em IEEE/ASME Transactions on Mechatronics}, 23(4):1907--1918,
  August 2018.

\bibitem{tognon_omnidirectional2018}
Marco Tognon and Antonio Franchi.
\newblock Omnidirectional {Aerial} {Vehicles} {With} {Unidirectional}
  {Thrusters}: {Theory}, {Optimal} {Design}, and {Control}.
\newblock {\em IEEE Robotics and Automation Letters}, 3(3):2277--2282.

\bibitem{ryll_novel2015}
Markus Ryll, Heinrich~H. Bülthoff, and Paolo~Robuffo Giordano.
\newblock A {Novel} {Overactuated} {Quadrotor} {Unmanned} {Aerial} {Vehicle}:
  {Modeling}, {Control}, and {Experimental} {Validation}.
\newblock {\em IEEE Transactions on Control Systems Technology},
  23(2):540--556, March 2015.

\bibitem{kamel_voliro2018}
Mina Kamel, Sebastian Verling, Omar Elkhatib, Christian Sprecher, Paula Wulkop,
  Zachary Taylor, Roland Siegwart, and Igor Gilitschenski.
\newblock The {Voliro} {Omniorientational} {Hexacopter}: {An} {Agile} and
  {Maneuverable} {Tiltable}-{Rotor} {Aerial} {Vehicle}.
\newblock {\em IEEE Robotics \& Automation Magazine}, 25(4):34--44, December
  2018.

\bibitem{allenspach_design_2020}
Mike Allenspach, Karen Bodie, Maximilian Brunner, Luca Rinsoz, Zachary Taylor,
  Mina Kamel, Roland Siegwart, and Juan Nieto.
\newblock Design and optimal control of a tiltrotor micro-aerial vehicle for
  efficient omnidirectional flight.
\newblock {\em The International Journal of Robotics Research},
  39(10-11):1305--1325, September 2020.
\newblock Publisher: SAGE Publications Ltd STM.

\bibitem{sal2022simultaneous}
Firat Sal.
\newblock Simultaneous swept anhedral helicopter blade tip shape and
  control-system design.
\newblock {\em Aircraft Engineering and Aerospace Technology}, 95(1):101--112,
  2022.

\bibitem{ryll_modeling_2016}
Markus Ryll, Davide Bicego, and Antonio Franchi.
\newblock Modeling and control of {FAST}-{Hex}: {A} fully-actuated by
  synchronized-tilting hexarotor.
\newblock In {\em 2016 {IEEE}/{RSJ} {International} {Conference} on
  {Intelligent} {Robots} and {Systems} ({IROS})}, pages 1689--1694, October
  2016.
\newblock ISSN: 2153-0866.

\bibitem{ryll_fast-hexmorphing_2022}
Markus Ryll, Davide Bicego, Mattia Giurato, Marco Lovera, and Antonio Franchi.
\newblock {FAST}-{Hex}—{A} {Morphing} {Hexarotor}: {Design}, {Mechanical}
  {Implementation}, {Control} and {Experimental} {Validation}.
\newblock {\em IEEE/ASME Transactions on Mechatronics}, 27(3):1244--1255, June
  2022.
\newblock Conference Name: IEEE/ASME Transactions on Mechatronics.

\bibitem{bicego2020nonlinear}
Davide Bicego, Jacopo Mazzetto, Ruggero Carli, Marcello Farina, and Antonio
  Franchi.
\newblock Nonlinear model predictive control with enhanced actuator model for
  multi-rotor aerial vehicles with generic designs.
\newblock {\em Journal of Intelligent \& Robotic Systems}, 100:1213--1247,
  2020.

\bibitem{brescianini_omni-directional_2018}
Dario Brescianini and Raffaello D’Andrea.
\newblock An omni-directional multirotor vehicle.
\newblock {\em Mechatronics}, 55:76--93, November 2018.

\bibitem{michieletto2018fundamental}
Giulia Michieletto, Markus Ryll, and Antonio Franchi.
\newblock Fundamental actuation properties of multirotors: Force--moment
  decoupling and fail--safe robustness.
\newblock {\em IEEE Transactions on Robotics}, 34(3):702--715, 2018.

\bibitem{lee_geometric_2010}
Taeyoung Lee, Melvis Leok, and N~Harris McClamroch.
\newblock Geometric tracking control of a quadrotor {UAV} on {SE}(3).
\newblock In {\em 49th {IEEE} conference on decision and control}, pages
  5420--5425. IEEE, 2010.

\bibitem{lin2001line}
Faa-Jeng Lin and Chih-Hong Lin.
\newblock On-line gain-tuning ip controller using rfnn.
\newblock {\em IEEE Transactions on Aerospace and Electronic Systems},
  37(2):655--670, 2001.

\bibitem{el2013application}
Aulia El~Hakim, Hilwadi Hindersah, and Estiko Rijanto.
\newblock Application of reinforcement learning on self-tuning pid controller
  for soccer robot multi-agent system.
\newblock In {\em 2013 joint international conference on rural information \&
  communication technology and electric-vehicle technology}, pages 1--6. IEEE,
  2013.

\end{thebibliography}
	\bibliographystyle{unsrt}	
\end{document}